\documentclass[accepted]{uai2024} % for initial submission
\usepackage[american]{babel}
\usepackage{booktabs}       % professional-quality tables
\usepackage{amsfonts}       % blackboard math symbols
\usepackage{nicefrac}       % compact symbols for 1/2, etc.
\usepackage{microtype}      % microtypography
\usepackage{setspace}
\usepackage{amsmath}
\usepackage{amssymb, bm}
\usepackage[dvipsnames,svgnames]{xcolor}
\usepackage{xcolor}
\usepackage{hyperref}
\usepackage{mathrsfs}

\usepackage{amsthm}
\usepackage[round]{natbib}

\usepackage[round]{natbib}
\setcitestyle{authoryear,open={(},close={)}} %Citation-related commands

\usepackage[utf8]{inputenc} % allow utf-8 input
\usepackage[T1]{fontenc}    % use 8-bit T1 fonts
\usepackage[]{hyperref}       % hyperlinks
\usepackage{booktabs}       % professional-quality tables
\usepackage{amsfonts}       % blackboard math symbols
\usepackage{nicefrac}       % compact symbols for 1/2, etc.
\usepackage{microtype}      % microtypography
\usepackage{nicefrac}
\usepackage{amssymb}
\usepackage{amsmath}
\usepackage{amsthm}
\usepackage{cleveref}
\usepackage{autonum}
\usepackage{MnSymbol} 

\usepackage{caption}
\usepackage{url}

\usepackage{breakurl}
\usepackage[breaklinks]{hyperref}
\usepackage{lettrine}
\usepackage{framed}
\usepackage{color}

\usepackage{algorithm,algpseudocode}

\algnewcommand{\algorithmicforeach}{\textbf{for each}}
\algdef{SE}[FOR]{ForEach}{EndForEach}[1]
  {\algorithmicforeach\ #1\ \algorithmicdo}% \ForEach{#1}
  {\algorithmicend\ \algorithmicforeach}% \EndForEach

\usepackage{dsfont}
\usepackage{booktabs} 
\usepackage{bm}
\usepackage{bbm}
\usepackage{ifthen}
\usepackage{graphicx}
\usepackage{adjustbox}
\usepackage{subcaption}
\usepackage{dsfont}
\usepackage{framed}
\usepackage{overpic}
\usepackage[svgnames]{xcolor}
\newtheorem{theorem}{Theorem}
\usepackage{apxproof}
\newtheorem{proposition}{Proposition}

\newtheorem{assumption}{Assumption}

\usepackage{xr}
\hypersetup{
	colorlinks=true,
	linkcolor=blue,
	urlcolor=magenta,
	citecolor=DarkBlue,
	linkbordercolor={0 0 1}
}

\newtheoremrep{theorem}{Theorem}
\renewcommand{\algorithmicrequire}{\textbf{Input:}}

\newcommand{\edit}[1]{{\color{red} (AS: #1)}}

\newcommand{\BQ}{\textsc{BanditQ} }

\theoremstyle{definition}

\title{\BQ: Fair Bandits with Guaranteed Rewards}

\author{Abhishek Sinha}
\affil{ School of Technology and Computer Science,
Tata Institute of Fundamental Research,
Mumbai 400005, India \newline
\texttt{abhishek.sinha@tifr.res.in}}
\begin{document}
\maketitle

\begin{abstract} \label{abstract}
Classic no-regret multi-armed bandit algorithms, including the Upper Confidence Bound (\textsc{UCB}), \textsc{Hedge}, and \textsc{EXP3}, are inherently unfair by design. Their unfairness stems from their  objective of playing the most rewarding arm as frequently as possible while ignoring the rest. In this paper, we consider a fair prediction problem in the stochastic setting with a guaranteed minimum rate of accrual of rewards for each arm. We study the problem in both full-information and bandit feedback settings. Combining queueing-theoretic techniques with adversarial bandits, we propose a new online policy, called \BQ, that achieves the target reward rates while conceding a regret and target rate violation penalty of at most $O(T^{\nicefrac{3}{4}}).$ The regret bound in the full-information setting can be further improved to $O(\sqrt{T})$ under either a monotonicity assumption or when considering time-averaged regret. The proposed policy is efficient and admits a black-box reduction from the fair prediction problem to the standard adversarial MAB problem. 
The analysis of the \BQ policy involves a new self-bounding inequality, which might be of independent interest. 
%
%\textcolor{red}{In our regret definition, we consider a somewhat weaker benchmark where the reward and constraint vectors are averaged over a window of constant size $w \geq 1$ instead over the entire horizon of length $T.$ }
	\end{abstract}

\section{Introduction}   \label{statement} 
A vast majority of the multi-armed bandit (MAB) algorithms deployed in practice are designed to maximize the cumulative rewards. Consequently, these algorithms could end up systematically avoiding a subset of arms (which could represent users with certain demographic characteristics or historical activities) the algorithm finds less rewarding \citep{sweeney2013discrimination}. In a typical case of algorithmic discrimination, Facebook was sued for targeting ads on housing, credit and employment by race, gender, and religion - all protected classes under US law \citep{hao2019facebook}. A similar problem of fair allocation of resources arises in wireless settings, where schedulers maximizing the total throughput could result in not serving a subset of users having relatively poor channels. 
%Since the rewards are stochastic, playing an arm may result in random rewards. 
A number of papers have proposed a solution to the fairness problem by putting an explicit constraint on the \emph{minimum frequency} of pulls for each arm. However, in many problems of practical interest, the algorithm designer is interested in guaranteeing a minimum rate of \emph{reward accrual} for each arm - not just ensuring a minimum frequency at which the arms ought to be pulled. 

\textbf{Examples:} (1) In online ad allocation, the advertisers are primarily interested in maximizing their click-through rate, which fetches them monetary rewards,  rather than just the number of times their ads are displayed against a search result. (2) In wireless scheduling problems, the users, who correspond to bandit arms in our formulation, are interested in guaranteed data rates rather than their frequency of scheduling - a low-level metric transparent to the users. 
%In online crowd-sourcing platforms, workers are primarily interested in the amount of money they make over a given span of time compared to the number of times they are assigned a job. 
(3) As a final example, consider a crowdsourcing platform (e.g., Amazon Mechanical Turk) where the workers receive payments for performing tasks \citep{fu2021fairness}. Upon completing each task, the platform receives a fixed percentage of the payment as revenue. The goals of the platform are - (a) to allocate the oncoming tasks fairly among the workers and (b) to maximize the platform's total revenue. In our formulation, the workers correspond to the arms, and the revenue maximization problem (b) becomes equivalent to the regret minimization problem. However, without the fairness requirement (a), the platform would assign most of the jobs to the best-performing workers, effectively ignoring a vast majority of the registered workers who may leave the platform dissatisfied. Hence, the platform may suffer from a high attrition rate. One possible way to enhance the retention rate of the workers and make the platform non-discriminating is to ensure a guaranteed reward rate (equivalent to a minimum wage) for each registered workers. In this paper, we will see that the proposed \textsc{BanditQ} policy gives an efficient solution to each of the above problems. 
%Similar problems also arise in other settings, including user scheduling in wireless networks where the network operator wants to ensure guaranteed bit rates to the users.

Clearly, the rate of reward accruals of the arms depends on the unknown reward distribution, which needs to be learned along the way. In this paper, we solve this fair prediction problem in the stochastic setting via a black-box reduction to an adversarial MAB problem by making use of a natural queueing dynamics to keep track of the target rates. Although we consider i.i.d. rewards, we will see that the use of adversarial MAB sub-routines is essential to account for the target reward rate constraints.   

\subsection{Related Works} \label{related}
There is an extensive literature on the classic Multi-armed Bandits (MAB) problem, where the objective is to sequentially play an arm on each round from a given set of arms with  unknown reward distributions to maximize the cumulative reward. As the feedback is limited to the observed rewards only, the MAB problem naturally involves an exploration vs exploitation trade-off. See \citet{cesa2006prediction, bubeck2012regret, lattimore2020bandit} for textbook treatments on MAB. The fair prediction problem considered in this paper belongs to a class of MAB problems with global constraints. Several authors have considered variants of the fair prediction problem in MAB with widely varying definitions for fairness \citep{joseph2016fairness, gillen2018online, bechavod2020metric, hossain2021fair, huang2022achieving}. Closer to our setting, the papers by \citet{patil2021achieving, claure2020multi}, and  \citet{li2019combinatorial} considered a stochastic MAB problem while requiring the minimum \emph{fraction} of pulls of each arm to exceed a given threshold. \citet{celis2019controlling} considered a similar problem in the personalized recommendation setting where both the minimum and the maximum fraction of pulls are constrained in order to avoid the polarization of views. Similar to ours, \citet{li2019combinatorial} used a virtual queueing recursion to handle the fairness constraints. However, their UCB-based policy yields a regret bound which varies \emph{linearly} with the horizon length \citep[Theorem 2]{li2019combinatorial}.  \citet{chen2020fair} considered the above problem in the contextual bandit setting and proposed a no-regret policy with a known context distribution. \citet{cai2018online} considered a related stochastic MAB problem with a long-term constraint on an auxiliary (level-$2$) reward process, which is assumed to be \emph{independent} of the main (level-$1$) rewards of the arms. On the other hand, in our problem, the corresponding level-$1$ and level-$2$ reward processes are identical, and hence, these results do not apply due to the lack of the independence assumption. \citet{badanidiyuru2018bandits, immorlica2022adversarial}, and  \citet{xia} considered the Bandits with Knapsack (BwK) problem in the stochastic and adversarial settings. In this problem, a given resource budget is allocated to the arms at the beginning, and the policy continues until one of the arms finishes all of its budgets. \citet{immorlica2022adversarial} used a Lagrangian-based technique to design a no-regret policy for the BwK problem. 
%Similar to ours, they also employed an adversarial bandit policy in the stochastic setting. 
A recent paper by \citet{bistritz2022queue} considered a similar multiplayer multi-armed bandit problem with QoS constraints. However, they did not provide any regret bound. In this connection, we also mention a parallel line of work on fair resource allocation policies where, instead of meeting explicit constraints, the objective is to maximize a non-linear concave utility function of the cumulative rewards \citep{nofra}. Our problem is also closely connected to a recent series of works on Online Convex Optimization (OCO) with long-term constraints \citep{neely2017online, yu2017online, yuan2018online, castiglioni2022unifying}. While these papers propose problem-specific policies, we give a black-box reduction using any arbitrary adaptive learning policy as a subroutine and achieve the state-of-the-art regret and constraint violation bounds. Furthermore, while most of the previous papers consider the full-information setting and/or assume the strict feasibility or Slater's condition, we consider the more general bandit feedback setting \emph{without} making any additional assumption. The Lyapunov-based technique presented in this paper has been recently extended to solve the problem of OCO with long-term constraints as well \citep{sinha2023playing}.

%The classic Proportional Fair scheduler solves the fair scheduling problem when the current channel conditions are known \citep{stolyar2005asymptotic}. However, maximizing the sum rate with arm-specific rate constraints - a central problem in 5G network slicing remains open under unknown channel conditions. In the context of online learning, 

%\cmt{mention the INFOCOM paper with $T^5/6$ regret}. 

%Broadly speaking, the fair prediction problem belongs to a class of reinforcement learning problem with global constraints. 
%Finally, we also mention a recent line of work that proposes no-regret algorithms for general linear dynamical systems which, in principle, can be used to model constrained MAB problems 
%\citep{hazan2017learning, hazan2018spectral}.  
\subsection{Our contributions}
In contrast with a major line of work on fair MABs, which is mainly concerned with guaranteeing a minimum frequency of plays for each arm (\emph{procedural fairness}), in this paper, we initiate the study of a class of problems guaranteeing a minimum \emph{rate} of reward accruals for each arm (\emph{substantive fairness}).
%, which depends on the \emph{unknown} reward distributions of the arms. 
Compared to the standard MAB problem, here, the difficulty stems from the fact that in addition to playing the unknown best arm sufficiently many times, other arms with unknown mean rewards also ought to be played frequently enough so as to satisfy the given fairness constraints. Consequently, the design of our algorithm and its analysis proceed along a different line from that of the prior works. In particular, we claim the following contributions:

\begin{enumerate}
	%\item We use adversarial experts algorithm for solving a constrained prediction problem in the stochastic setting. The technical reason for this approach is that, the constrained problem is reduced to an instance of unconstrained problem with a new set of rewards which could be correlated in a complex fashion.
	\item We propose a fair learning policy for stochastic bandits, called \textsc{BanditQ}, via a \emph{black-box} reduction to the standard adversarial MAB problem. The problem is studied in both full information and bandit feedback settings. The proposed \BQ policy keeps track of the global reward rate constraints through an auxiliary queueing process, which is then used to define the rewards for the unconstrained MAB problem recursively. 
	
	\item An attractive feature of our policy is that it is completely oblivious to the algorithm used for the unconstrained MAB problem. In particular, \BQ can use \emph{any} existing MAB policy with a data-dependent adaptive regret bound. The key to this attractive separation result is a new \emph{self-bounding} inequality that bounds the sum of the regret and current rate violations in terms of the past violations. 
	%This decomposition technique could be useful for studying other constrained sequential learning problems as well.
	\item We introduce a new proof technique that bounds the regret and rate violations by solving certain sequential inequalities. The proof arguments are crisp and utilize off-the-shelf adaptive regret bounds.
	\item We supplement our theoretical results with illustrative numerical experiments.
	%	\item We give a unified treatment for the full-information and bandit-feedback cases with almost identical proofs. 
	\end{enumerate}
 %As a technical note, compared to previous works on bandits with knapsacks where two different online policies are used (a primal and a dual \citep{immorlica2022adversarial}), in this paper, we use only one explicit no-regret policy as a subroutine. This makes our algorithm more efficient.
	%The key to our results is the use of adversarial MAB policies for handling the stochastic problem. 
	
%A similar approach was adopted by \citet{immorlica2022adversarial} where they used nearly identical policy for solving the stochastic and adversarial Bandits with Knapsacks (\textsc{BwK}) problem with optimal regret bounds. 
%\end{itemize}
  
%, which can be trivially solved in our setting by a randomized policy. 

%We consider the standard no-regret prediction problem with expert's advice in the adversarial setting with an additional long-term rate constraint for a subset of experts. 

%\textcolor{red}{Replace agents by arms?}

%More specifically, we use the strong $(w, \epsilon)$-admissibility criterion with respect to a fixed schedule. The problem is interesting both in the full information and bandit setting and we tackle the full information version first. 

%We will also consider the simpler problem where there is a  given rate vector $\bm{r}$ such that arm $i$ must be pulled at least $r_i$ fraction of times. We seek a uniform algorithmic framework for both these problems. 

\section{Problem formulation}  \label{model}  
We consider a regret minimization problem in the context of Multi-armed Bandits (MAB) with an additional fairness constraint. The fairness constraint requires that each arm in a given subset $\mathcal{P}$ (called \emph{protected class}) must attain pre-specified reward accrual rates, which are assumed to be feasible. Formally, we consider an $N$-armed bandit, which on round $t$ receives an unknown reward vector $\bm{r}(t) \in [0,1]^N.$ The vector $\bm{r}(t)$ is generated i.i.d. on each round with an unknown expectation $\bm{\mu}.$ On round $t$, an online policy first decides a  probability distribution $\bm{x}(t) \in \Delta_N,$ where $\Delta_N$ denotes the set of all probability distributions supported on $N$ arms. The policy then randomly samples an arm $I_t \in [N]$ from the distribution $\bm{x}(t)$\footnote{This protocol includes conditionally deterministic policies, such as \textsc{UCB}, where $\bm{x}(t)$ is supported on only one arm.}. Depending on the feedback structure, either the entire reward vector $\bm{r}(t)$  (in the case of full information feedback) or the reward of the sampled arm $r_{I_t}(t)$ only (in the case of bandit feedback) is revealed to the policy at the end of round $t$. The above process continues for $T$ rounds.   
%Due to the linearity of the rewards, it is sufficient for the policy to output a sequence of distributions $\{\bm{x}(t)\}_{t\geq 1}.$
%This short paper focuses on the full-information setting where the entire reward vector is revealed to the learner at the end of each round. 
\paragraph{Fairness constraints:}
%If an arm $I_t$ is selected by the policy on round $t$, then this arm 
Due to the action of the policy, the selected arm $I_t$ receives a random reward of value $r_{I_t}(t).$ 
Hence, if on round $t$, the online policy samples arms according to the distribution $\bm{x}(t)$, the $i$\textsuperscript{th} arm receives a (conditional) expected reward of $x_i(t)\mathbb{E}r_i(t)=x_i(t)\mu_i,$ and the online policy receives an overall (conditional) expected reward of $\langle \bm{x}(t), \bm{\mu} \rangle.$ 
%Let $\mathcal{P} \subseteq [N]$ denote the subset of arms belonging to the \emph{protected classes}. 
Let $\bm{\vec{\lambda}}$ be the given target reward rates vector.
The fairness constraint mandates that the long-term rate of rewards accrued by arm $i \in \mathcal{P},$ must be at least $\lambda_i, \forall i\in \mathcal{P}$ (see Eqn.\ \eqref{rate-constr2}). For notational simplicity, we may assume that $\lambda_i=0, \forall i \in [N]\setminus \mathcal{P}.$ 
%We emphasize that guaranteeing lower bounds to the rate of reward accruals of the individual arms is different and more challenging than just constraining the frequency of sampling each arm without considering their accrued rewards. 
%To make the rate constraints feasible, 
%we assume that there is an offline probability distribution $\bm{x}^*$ over the arms with the above property. 
%we assume that there exists a feasible offline static allocation 
\paragraph{Offline Benchmark and Performance Metric:}
We compare the performance of an online policy against any fixed sampling distribution
$\bm{x}^* \in \Delta_N$ that meets the target reward rates. In other words, our comparator class $\Omega(\bm{\vec{\lambda}})$, indexed by the target vector $\bm{\vec{\lambda}}$, is defined as follows:
\begin{eqnarray} \label{feas}
	\Omega(\bm{\vec{\lambda}}) = \big\{\bm{x}^* \subseteq \Delta_N: x_i^* \mu_i \geq \lambda_i, ~\forall i \in \mathcal{P}\big\}.
 %\nonumber\\
%	&&\textrm{and}~ \sum_i x_i^* =1, ~\bm{x}^* \geq \bm{0}\big\}.
\end{eqnarray}  
Clearly, in order for the target rate vector $\bm{\vec{\lambda}}$ to be feasible (\emph{i.e.,} $\Omega(\bm{\vec{\lambda}}) \neq \emptyset$) it is necessary and sufficient that 
\begin{eqnarray} \label{feas-constr}
	\sum_i \frac{\lambda_i}{\mu_i} \leq 1.
\end{eqnarray}
See Section \ref{feas-sec} in the Appendix for a brief discussion on the feasibility assumption. 
The set of all offline benchmarks $\Omega(\bm{\vec{\lambda}})$ is closed and convex with an Euclidean diameter of $D=\sqrt{2}.$ Our goal is to design a sampling policy $\{\bm{x}(t)\}_{t\geq 1}$ that achieves a sublinear regret against any $\bm{x}^* \in \Omega(\bm{\vec{\lambda}}),$ where 
%we use the standard definition of (pseudo-) regret:
\begin{eqnarray} \label{regret_def}
	\textrm{Regret}_T(\bm{x}^*) \equiv 
	%\max_{\bm{x}^* \in \Omega(\bm{\vec{\lambda}})} 
	\langle \bm{x}^*, \bm{\mu}\rangle T - \mathbb{E}\sum_{t=1}^T \sum_{i=1}^N r_i(t)\mathds{1}(I_t=i),
\end{eqnarray}
while meeting the long-term reward rate constraints defined next\footnote{When we refer to the worst-case regret, we drop the argument $\bm{x}^*$ in parenthesis in the regret definition \eqref{regret_def}.}. Asymptotically, for any time interval $\mathcal{I} \subseteq [T]$, the long-term rate constraint requires:
\begin{eqnarray} \label{rate-constr2}
	\liminf_{|\mathcal{I}| \to \infty} |\mathcal{I}|^{-1}\mathbb{E}\big[\sum_{t \in \mathcal{I}} r_i(t)\mathds{1}(I_t=i)\big]  \geq \lambda_i, ~\forall i \in \mathcal{P}.  
\end{eqnarray}
Note that Eq.\ \eqref{rate-constr2} requires the minimum reward rate guarantee to hold \emph{uniformly} across the time horizon for any sufficiently long interval of time. In other words, we require that no individual arm is starved for a long period of time - a problem left open by \citet{patil2021achieving}. Furthermore, following \citet{cai2018online}, we work with a fine-grained non-asymptotic metric \emph{rate violation penalty} defined below:
\begin{eqnarray} \label{violation_penalty}
	\mathbb{V}(T) = \max_{i \in P} \mathbb{E}\big[ \sum_{t=1}^T \big(\lambda_i - r_i(t) \mathds{1}(I_t=i)\big) \big].
\end{eqnarray} 
In brief, we seek to design an online policy for which \emph{both} $\textrm{Regret}_T$ and $\mathbb{V}(T)$ are sub-linear in $T$.
%In this paper, we will give a stronger \emph{uniform} guarantee of the reward rate achieved by our proposed policy.
We note two fundamental differences between the above problem and the standard online learning framework \citep{orabona2019modern}. First, contrary to the online learning setting, where the set of benchmarks $\Omega$ is specified a priori (independent of the rewards), in this problem, the set of benchmarks \eqref{feas} depends on the unknown reward distributions through their expectations $\bm{\mu}$. Second, unlike the online learning setting, the action taken by the policy on a round is not restricted to the set $\Omega(\bm{\vec{\lambda}})$ provided that the long-term target rates are met. Note that upon setting the vector $\vec{\bm{\lambda}}$ to zero, we recover the classic MAB problem as a special case. 
%In Appendix \ref{adapt}, we give a detailed discussion on the difficulty of adapting well-known bandit policies to this problem. 
In the following section, we introduce the \BQ policy in the full information setting.
\section{\large{\BQ} Policy with full information feedback}  \label{algorithm}
%In this section, we propose \textsc{BanditQ} - an online learning policy that solves the above constrained prediction problem. 
%Although our major result pertains to the bandit feedback setting, 
%for simplicity of exposition, we first 
In this Section, we consider the full-information setup when the entire reward vector is revealed to the learner at the end of each round. Apart from a technical result regarding the diameter of an auxiliary random process (Proposition \ref{uniform_bd_lemma} in the Appendix), the extension of the full-information policy to the bandit setting requires no substantially new ideas and will be dealt with in the following section. 
On a high level, the \textsc{BanditQ} policy first defines a \emph{queueing} dynamics to take into account the gap between target reward and the reward accrued by the policy for each arm so far. It then extends the \emph{drift-plus-penalty} framework of \citet[Chapter 4]{neely2010stochastic} to simultaneously achieve a small regret and meet the long-term constraints. However, to make this high-level scheme work, we must adapt the asymptotic stochastic setting of \cite{neely2010stochastic} to the non-asymptotic adversarial setup with online information. This extension turns out to be highly non-trivial and requires new proof and algorithmic techniques, which are very different from that of the \textsf{Max-Weight} policy proposed by \citet{neely2010stochastic}.

We associate a non-negative state variable $Q_i(t)$ to each protected arm $i \in \mathcal{P}.$ Under the action of an online policy $\pi = \{\bm{x}(t)\}_{t \geq 1},$ the state variables evolve according to the following queueing dynamics, known as the Lindley recursion \citep{lindley1952theory}:
\begin{eqnarray} \label{q-ev}
	Q_i(t)=\big(Q_i(t-1)+ \lambda_i - r_i(t)x_i(t)\big)^+, ~Q_i(0)=0,
\end{eqnarray}
where we adopt the standard notation $(y)^+\equiv \max(0,y).$  We set $Q_i(t)=0, \forall t, \forall i \notin \mathcal{P}$. To get an intuition for Eq.\  \eqref{q-ev}, imagine that on every round $t,$ a fixed deterministic amount of work $\lambda_i$ arrives at the queue $Q_i.$ Then, under the action $\bm{x}(t)$ of an online policy, $\min(Q(t-1)+ \lambda_i, r_i(t) x_i(t))$ amount of work departs from $Q_i.$ It is intuitive that to stabilize the queues, the long-term service rates must be at least as large as the long-term arrival rates. Thus, any online policy stabilizing the queues would automatically satisfy the target rate requirements. However, since we are also interested in achieving a small regret, meeting the rate constraints alone is not enough (\emph{c.f.} \cite{huang2023queue}). Our online policy must also perform competitively in terms of the cumulative rewards against every feasible stationary action given by \eqref{feas}. 

Towards this goal, let us first define the following quadratic potential function (\emph{a.k.a.} Lyapunov function in the queueing theory parlance):
\begin{eqnarray} \label{potential_def}
	\Phi(t) = \sum_{i \in \mathcal{P}} Q_i^2(t).  
\end{eqnarray}
We now upper bound the change of the potential under the action of a policy. From \eqref{q-ev}, we have 
\begin{eqnarray*}
	&&Q_i^2(t) \\
	 &\leq& \big(Q_i(t-1)+ \lambda_i - r_i(t)x_i(t)\big)^2 \\
	&\leq& Q_i^2(t-1) + \lambda_i + x_i(t)+ 2Q_i(t-1)(\lambda_i - r_i(t) x_i(t)), 
\end{eqnarray*}
where, in the last inequality, we have used the fact that $0\leq \lambda_i, r_i(t), x_i(t) \leq 1, \forall i, t.$ Summing up the above inequality for each $i \in \mathcal{P}$, we have the following upper bound for the change of the potential on round $t$:
\begin{eqnarray} \label{potential_change}
	\Phi(t) - \Phi(t-1) \leq 2 + 2\sum_{i \in \mathcal{P}} Q_i(t-1)\big(\lambda_i - r_i(t) x_i(t)\big),
\end{eqnarray}
where we have used the fact that $\sum_i \lambda_i \leq 1, \sum_i x_i(t) \leq 1,$ where the first inequality follows from the non-emptiness of $\Omega(\bm{\vec{\lambda}}).$  Eqn.\ \eqref{potential_change} suggests that running a MAB policy to maximize virtual cumulative rewards such that pulling the $i$\textsuperscript{th} arm on round $t$ yields a virtual reward of $Q_i(t-1)r_i(t)$ will help minimize the change of the potential on round $t$ and, hence, meet the target rates. However, this does not explicitly take into account our other goal, namely to minimize the regret. To achieve both the goals, motivated by the drift-plus-penalty framework of \cite{neely2010stochastic}, we now define an instance of the standard online linear optimization (OLO) problem $\Xi$ with action set $\Delta_N$, where the surrogate reward of the $i$\textsuperscript{th} arm on round $t$ is defined as: 
\begin{eqnarray} \label{reward_def}
	r'_i(t) \equiv \big(Q_i(t-1) + V\big)r_i(t), ~\forall i \in [N].
\end{eqnarray}
In the above, $V>0$ is a hyper-parameter, to be fixed later, that depends only on the length of the horizon $T.$
 %$\{V_t\}_{t\geq 1}$ is a sequence of non-negative parameters. In our theoretical results, we will primarily consider a constant sequence $V_t=V=\Theta(\sqrt{T}), \forall t.$ 
 Intuitively, the surrogate reward vector $\bm{r}'(t)$ strikes a balance between attaining the target rates (through the first term) and achieving a small regret (through the second term). However, the reward definition \eqref{reward_def} leads to two significant technical challenges for learning the surrogate rewards. First, due to the presence of the queue variables, the reward vectors $\bm{r}'(t)$ are not bounded \emph{a priori}, which critically affects the regret bound for the surrogate problem $\Xi$. Second, although the original reward sequence $\{\bm{r}(t)\}_{t\geq 1}$ is i.i.d., the reward sequence $\{\bm{r}'(t)\}_{t \geq 1}$ for the problem $\Xi$ is \emph{not} i.i.d. any more, again due to the presence of the queue variables, which are temporally correlated via Eq.\ \eqref{q-ev}. The second difficulty prompts us to use an adversarial online learning policy for the auxiliary OLO problem $\Xi.$ 
 %To recapitulate the information structure, the coefficients $Q_i(t-1)+V_t$ are known, but the reward vector $\bm{r}(t)$ is unknown to the online policy before it makes the decision $\bm{x}(t) \in \Delta_N$ on round $t$. 
 \paragraph{The \BQ policy:} 
 The proposed \textsc{BanditQ} policy can use any adaptive no-regret policy with a second-order regret bound for the auxiliary problem $\Xi.$ This includes policies such as Online Gradient Ascent (\textsc{OGA}) with adaptive step sizes \citep{orabona2019modern} \footnote{Since ours is a maximization problem, we use a gradient ascent step rather than descent.} and \textsc{Squint} \citep{koolen2015second}. To fix ideas, in this paper, we will use the \textsc{OGA} policy due to its simplicity. This online policy, which is closely related to the AdaGrad policy \citep{duchi2011adaptive}, updates the sampling distribution on each round using the usual gradient step with an adaptive step size:
\begin{eqnarray} \label{oga-update}
	\bm{x}(t+1) \gets \Pi_{\Delta_N}\bigg(\bm{x}(t) + \frac{\bm{r}'(t)}{\sqrt{2 \sum_{\tau=1}^{t}||\bm{r}'(\tau)||_2^2}} \bigg).
\end{eqnarray}
%where the reward sequence $\{\bm{r}'(t)\}_{t\geq 1}$ is defined in Eq.\ \eqref{reward_def}.
%, and the state variables $\{\bm{Q}(t)\}_{t \geq 1}$ evolve as in Eq. \eqref{q-ev}. 
In the above, the $\Pi_{\Delta_N}(\cdot)$ function, which denotes the Euclidean projection operator on the standard simplex $\Delta_N,$ can be efficiently implemented in $O(N\log N)$ time \citep{wang2013projection}. The complete \BQ policy in the full-information setting is summarized in Algorithm \ref{fair-MAB-full-info}.

  \begin{algorithm}
\caption{\BQ Policy with full information}
\label{fair-MAB-full-info}
\begin{algorithmic}[1]
\State \algorithmicrequire{ Target reward rate vector $\bm{\vec{\lambda}}$,} Euclidean projection oracle $\Pi_{\Delta_N}(\cdot)$ onto the simplex $\Delta_N.$ 
\State $\bm{Q} \gets \bm{0}, \bm{x} \gets [\nicefrac{1}{N}, \nicefrac{1}{N}, \ldots, \nicefrac{1}{N}], V\gets \sqrt{T}, S\gets 0.$ \algorithmiccomment{\emph{Initialization}}
\ForEach {round $t=1:T$}
\State Sample an arm $I_t$ from the distribution $\bm{x}$.  
\State Observe the \emph{entire} reward vector $\bm{r}(t)$\algorithmiccomment{\emph{Full-information feedback}}
%\ForEach {arms $i \in \mathcal{P}$:}
\State 
%\begin{eqnarray*} 
	$Q_i=\big(Q_i+ \lambda_i - r_i(t)x_i\big)^+, ~\forall i\in \mathcal{P}. $\algorithmiccomment{\emph{Updating the queue lengths}}
%\end{eqnarray*}
%\EndForEach
\State $r'_i(t) \gets \big(Q_i + V\big)r_i(t), ~\forall i \in [N]$ \algorithmiccomment{\emph{Computing the surrogate rewards}}
\State $S \gets S + ||\bm{r}'(t)||^2.$ \algorithmiccomment{\emph{Accumulating the norm of the past gradients}}
	\State $\bm{x}\gets \Pi_{\Delta_N}\bigg(\bm{x}+ \frac{\bm{r}'(t)}{\sqrt{2S}} \bigg)$ \algorithmiccomment{\emph{Online gradient ascent}}
\EndForEach
\end{algorithmic}
\end{algorithm}
In our analysis, we will use the following standard second-order regret bound achieved by the \textsc{OGA} policy with the above adaptive step sizes. 
% \begin{framed} 
\begin{theorem}[\citet{orabona2019modern}, Theorem 4.14] \label{ref_th}
	Let $X \subseteq \mathbb{R}^{d}$ be a convex set with a finite Euclidean diameter $D.$ Consider an arbitrary sequence of linear reward functions with gradients $\{\bm{g}_t\}_{t \geq 1}.$ Assume that the Online Gradient Ascent policy is run with step sizes\footnote{Without any loss of generality, we set $\eta_t=0$ if $\bm{g}_t=0.$} $\eta_t= \frac{D}{\sqrt{2\sum_{\tau=1}^t|| \bm{g}_\tau||_2^2}}, 1\leq t \leq T.$ Then the regret of the policy can be upper-bounded as follows:
	\begin{eqnarray} \label{data-dep-bd}
	 \textrm{Regret}_T \leq D \sqrt{2\sum_{t=1}^T||\bm{g}_t||_2^2}. 		
	\end{eqnarray}
\end{theorem}  
%\end{framed}
 %for the problem $\Xi,$
 % which is defined by replacing the reward $\bm{r}(t)$ with $\bm{r}'(t)$ in Eq.\ \eqref{regret_def}:
 %A useful property of the above regret bound, which we exploit in our analysis, is that 
 It is important to note that the above bound is \emph{scale-free}, \emph{i.e.,} no \emph{a priori} bounds on the gradients are needed for the above result \citep{putta2022scale, hadiji2023adaptation}. Specializing Theorem \ref{ref_th} to our surrogate problem $\Xi$, we obtain the following regret bound which depends on the sequence of queue variables:
\begin{eqnarray} \label{regret-bound}
	\textrm{Regret}^\Xi_t &\leq& 2 \sqrt{\sum_{\tau=1}^t \sum_i (Q_i(\tau-1)+V)^2r_i(t)^2} \nonumber\\
	&\leq&  2\sqrt{2\sum_{\tau=1}^t\sum_iQ_i^2(\tau) } + 2V \sqrt{2Nt}.
\end{eqnarray}
  In the above, we have used the fact that $0\leq r_{i}(t)\leq 1, \forall t,i,$ and the elementary inequalities $(a+b)^2 \leq 2(a^2+b^2),$ $\sqrt{x+y} \leq \sqrt{x}+ \sqrt{y}, x,y\geq 0$.

\subsection{Analysis} \label{analysis}
Unlike the analysis in \citet{patil2021achieving} and \cite{cai2018online}, which proceed by constructing stochastic confidence intervals for the mean rewards of each arms, we directly make use of the regret bound \eqref{regret-bound} via an "adversarial-style" analysis, which critically makes use of a new \emph{self-bounding} inequality derived below. Since the state variables $\{\bm{Q}(t)\}_{t \geq 1}$ %corresponding to the arms in class $\mathcal{P}$ 
evolve according to the recursion \eqref{q-ev}, we do not immediately have an explicit control on the regret bound \eqref{regret-bound} which depends on the queue lengths. Hence, to bound the regret, we take an indirect approach. Fix any feasible distribution $\bm{x}^* \in \Omega$.
From Eq. \eqref{potential_change}, we have
\begin{eqnarray*}
&&\Phi(\tau)-\Phi(\tau-1)  - 2V \sum_i r_i(\tau)x_i(\tau) \\
&\leq& 2+ 2 \sum_i Q_i(\tau-1) \lambda_i -\\
&&2 \sum_{i} \underbrace{(Q_i(\tau-1)+V) r_i(\tau)}_{r_i'(\tau)} x_i(\tau).
\end{eqnarray*}
%Taking conditional expectation of both sides with respect to the randomness of the reward vector $\bm{r}(\tau)$, we have 
%\begin{eqnarray*}
%	d
%\end{eqnarray*}
%
%Changing the dummy variable from $t$ to $\tau$ and 
Summing up the above inequalities from $\tau=1$ to $\tau=t$ and recalling that $\Phi(t)=\sum_i Q_i^2(t), \Phi(0)=0,$ we obtain
%from $\tau=1$ to $\tau=t$, 
%we obtain: 
\begin{eqnarray} \label{main_ineq}
	&&\sum_{i} Q_i^2(t) + 2 \sum_{\tau=1}^t V \sum_i r_i(\tau) (x_i^*-x_i(\tau)) \nonumber \\
	 &\leq& 2t + 2 \sum_{\tau=1}^t \sum_{i} Q_i(\tau-1) \big(\lambda_i-r_i(\tau) x_i^* \big)+2\textrm{Regret}^{\Xi}_t,\nonumber \\
	%&&+ \textrm{Regret}^{\Xi}_t \nonumber \\
	%&&\leq  2t +  4 \sqrt{2\sum_{\tau=1}^t\sum_{i} Q_i^2(\tau) } + 4\sqrt{2N\sum_{\tau=1}^t V_\tau^2}, \\
\end{eqnarray} 
where $\textrm{Regret}^{\Xi}_t$ denotes the worst-case regret for the surrogate problem (defined similarly as Eq.\ \eqref{regret_def}). 
Note that, in the above, the regret bound on the RHS is random as it depends on the magnitude of the random process $\{\bm{Q}(\tau)\}_{\tau}$. 
%In our analysis, we will exclusively consider a constant $\{V_t\}_{t\geq 1}$ sequence, where $V_t=V, \forall t\geq 1$ for some appropriate $V \geq 0$ to be fixed later.
Let $\{\mathcal{F}_\tau\}_{\tau \geq 0}$ be the natural filtration generated by the sequence of rewards $\{\bm{r}(\tau)\}_{\tau \geq 0}.$ Taking expectations, we have the following set of inequalities for any benchmark distribution $\bm{x}^* \in \Omega(\bm{\vec{\lambda}})$:
%Taking expectation of both sides with respect to the randomness of the reward process, we have:
\begin{eqnarray} \label{main-ineq3}
&&\sum_{i} \mathbb{E}Q_i^2(t) + 2V \textrm{Regret}_t (\bm{x}^*) \nonumber \\
		&=&\sum_{i} \mathbb{E}Q_i^2(t) + 2V \sum_{\tau=1}^t  \mathbb{E}\sum_i r_i(\tau) (x_i^*-x_i(\tau)) \nonumber \\
		 &\stackrel{(a)}{\leq} & 2t + 2 \sum_{\tau=1}^t \mathbb{E}\sum_{i} Q_i(\tau-1) \big(\lambda_i-x_i^*\mathbb{E}[r_i(\tau) |\mathcal{F}_{\tau-1}] \big)+ \nonumber \\
		 &&2 \mathbb{E} \big[\textrm{Regret}^{\Xi}_t\big]\nonumber\\
	 &\stackrel{(b)}{\leq}& 2t + 2 \sum_{\tau=1}^t \mathbb{E}\sum_{i} Q_i(\tau-1) \big(\lambda_i-\mu_i x_i^* \big)+2 \mathbb{E}\big[\textrm{Regret}^{\Xi}_t\big]\nonumber\\
	 &\stackrel{(c)}{\leq} & 2t + 2 \mathbb{E}\big[ \textrm{Regret}^{\Xi}_t\big]\nonumber\\
	&\stackrel{(d)}{\leq} & 2t +  4\sqrt{2\sum_{\tau=1}^t\sum_i \mathbb{E}Q_i^2(\tau) } + 4V \sqrt{2Nt} ,
\end{eqnarray}
where in (a), we have taken the expectation of both sides of \eqref{main_ineq} with respect to the i.i.d.\ reward process $\{\bm{r}(t)\}_{t \geq 1},$ and used the law of iterated expectations; in (b) we have used the i.i.d.\ nature of the reward process; in (c) we have used the feasibility condition of the benchmark $\bm{x}^*$ from Eq.\ \eqref{feas}; in (d) we have used the second-order regret bound from Eq.\ \eqref{regret-bound} in conjunction with Jensen's inequality for the square-root function. We emphasize that step $(d)$ is the \emph{only} place where we use any property of the online learning subroutine. In other words, our reduction is \emph{universal} in the sense that any online learning subroutine for $\Xi$, which could be very different from OGA but has a data-dependent regret bound similar to \eqref{data-dep-bd}, can be used with \BQ.

 Inequality \eqref{main-ineq3} constitutes the key step in our analysis. It shows that the queue-length process $\{\bm{Q}(t)\}_{t \geq 1}$ possesses a \emph{self-bounding} property in the sense that the expected queue-length squared at any round $t$ is bounded by the square root of the sum of expected queue-length squared up to round $t$ plus other auxiliary terms. %Inequality \eqref{main-ineq3} leads to the following bound on the second moments of the queue variables.
% $\{\bm{Q}(t)\}_{t \geq 1}$. 
The regret decomposition inequality \eqref{main-ineq3} will be used to prove our main result in the full information setting.
%for a specific setting of the $\{V_t\}_{t \geq 1}$ sequence. 
\begin{theorem}\label{q_bd}
	%Upon setting $V=\Theta(\sqrt{T}), \forall t\geq 1$, 
	The \BQ policy described in Algorithm \ref{fair-MAB-full-info} achieves the following regret and rate violation bounds:
	\begin{eqnarray*}
		\textrm{Regret}_T = O(\max(\frac{T}{\sqrt{V}}, \sqrt{NT})), \mathbb{V}(T) = O(\sqrt{VT}).
	\end{eqnarray*}
	In particular, upon setting $V=\sqrt{T},$ we obtain 
		\begin{eqnarray*}
		\textrm{Regret}_T = O(\max(T^{\nicefrac{3}{4}},\sqrt{NT})), ~ \mathbb{V}(T) = O(T^{\nicefrac{3}{4}}).
	\end{eqnarray*}
		%we obtain $\mathbb{E}Q^2_i(t) = O(\sqrt{N}T^{3/2}), 1\leq t\leq T.$ 
\end{theorem} 
%\cmt{Improve the regret bound.}
The proof, given below, involves solving a non-linear sequential inequality to obtain a sublinear bound for the queue lengths. The resulting queue length bound is then used to control the regret. 
\begin{proof}
	First, we will derive a sublinear bound for the expected queue lengths under the \BQ policy. The rate violation and regret bounds will follow from this result.
\paragraph{1 (a). Bounding the queue lengths:}
Since the reward components are bounded in $[0,1],$ using the fact that $\sum_i r_i(\tau)(x_i(\tau)-x_i^*) \leq 1, \forall \tau,$ we have that $\textrm{Regret}_t(\bm{x}^*)\geq -t.$ Hence, from Eq.\ \eqref{main-ineq3}, we have for all $t\geq 1:$
\begin{eqnarray} \label{ineq1}
	\sum_i\mathbb{E}Q_i^2(t) \leq  2(V+1)t + 4 \sqrt{2\sum_{\tau=1}^t\sum_{i}\mathbb{E}Q_i^2(\tau) } \nonumber \\ + 4V \sqrt{2Nt}.
\end{eqnarray}
Hence, for any round $1\leq \tau \leq t,$ we have that 
\begin{eqnarray*}
		\sum_i \mathbb{E}Q_i^2(\tau) \leq 2(V+1)t + 4 \sqrt{2\sum_{\tau=1}^t\sum_{i}\mathbb{E}Q_i^2(\tau) } + 4V \sqrt{2Nt}.
\end{eqnarray*}
Summing up the above inequalities for all $\tau \in [1,t],$ we have 
\begin{eqnarray*}
	R^2(t) \leq 2(V+1)t^2 + 4 \sqrt{2N}V t^{\nicefrac{3}{2}} + 4\sqrt{2} t R(t).
\end{eqnarray*}
where we have defined $R(t) \equiv \sqrt{\sum_{\tau=1}^t \sum_{i=1}^N \mathbb{E}Q_i^2(\tau)}.$ Solving the above quadratic inequality in $R(t)$, we obtain
\begin{eqnarray} \label{r-bd}
	 R(t) = O(t)+ O(t\sqrt{V})+O(N^{\nicefrac{1}{4}}\sqrt{V}t^{\nicefrac{3}{4}})= O(t\sqrt{V}).
\end{eqnarray} 
Plugging the above bound in \eqref{ineq1}, we have for each $i \in \mathcal{P}:$
\begin{eqnarray} \label{q-sq-bd}
 &&\mathbb{E}Q_i^2(t) = O(Vt) + O(t\sqrt{V})+ O(V\sqrt{Nt}) = O(Vt) \nonumber \\
 && \stackrel{\textrm{(Jensen's ineq.)}}{\implies}  \mathbb{E}Q_i(t) = O(\sqrt{Vt}).	
\end{eqnarray}
\paragraph{1 (b). Bounding the rate violation penalty $\mathbb{V}(T)$:}
	Upon expanding \eqref{q-ev}, we obtain the following well-known representation for the Lindley recursion \citep[pp. 92]{asmussen2003applied}: 
	\begin{eqnarray} \label{q-len-bd}
		Q_i(t) = \sup_{1\leq \tau \leq t}(0, \lambda_i \tau - \sum_{z=t-\tau+1}^t r_i(z) x_i(z)), ~ \forall i \in \mathcal{P}.
	\end{eqnarray}
Combining Eq.\ \eqref{q-len-bd} with the bound \eqref{q-sq-bd}, we can bound the constraint violation penalty as 
	\begin{eqnarray*}
		\mathbb{V}(T) \leq \max_{i \in \mathcal{P}} \mathbb{E}Q_i(T) = O(\sqrt{VT}). 
	\end{eqnarray*} 
\paragraph{2. Bounding the regret:}
	Substituting \eqref{r-bd} into the inequality \eqref{main-ineq3} and using the fact that $Q_i^2(T) \geq 0, \forall i, t,$ we have for any $\bm{x}^* \in \Omega:$ 
	\begin{eqnarray*}
	2V \textrm{Regret}_T (\bm{x}^*)  
		\leq O(T) + O(T\sqrt{V})+ O(V\sqrt{NT}).
	\end{eqnarray*}
	This yields the following regret bound 
	\begin{eqnarray*}
		\textrm{Regret}_T(\bm{x}^*)  &=& O(\frac{T}{V})+ O(\frac{T}{\sqrt{V}})+ O(\sqrt{NT})\\
		&=& O(\max(\frac{T}{\sqrt{V}}, \sqrt{NT})).
	\end{eqnarray*}
\end{proof}

%\begin{proof}
%From Eq.\ \eqref{main-ineq3}, we have that for all $i \in \mathcal{P}$ and all $t\geq 1:$
%\begin{eqnarray} \label{ineq1}
%	\mathbb{E}Q_i^2(t) \leq 2(V+1)t + 4 \sqrt{2\sum_{\tau=1}^t\sum_{i}\mathbb{E}Q_i^2(\tau) } + 4V \sqrt{2Nt},
%\end{eqnarray}
%where we have used the fact that $\sum_i r_i(\tau)(x_i(\tau)-x_i^*) \leq 1, \forall \tau.$ Which implies that $\textrm{Regret}_t(\bm{x}^*)\geq -t.$ Furthermore, since $\lambda_i \leq 1,$ from Eq.\ \eqref{q-ev}, we trivially have $Q_i(\tau) \stackrel{a.s.}{\leq} \tau, \forall i,\tau$ We now improve upon this trivial upper bound on the queue lengths by substituting it on the RHS of Eq.\ \eqref{ineq1}, which yields:
%\begin{eqnarray*}
%	\mathbb{E}Q_i^2(t) &\leq& O(T^{3/2}) + O\big(\sqrt{N \sum_{\tau=1}^t \tau^2}\big)+ O(\sqrt{N} T) \\
%	&=& O(\sqrt{N} T^{3/2}),  \forall i, t \in [T].
%\end{eqnarray*}
%\end{proof}

\paragraph{Remarks:} It may appear from the statement of Theorem \ref{q_bd} that \BQ achieves a sub-optimal $O(T^{3/4})$ regret bound even for the standard regret minimization problem with no specified target reward rates, \emph{i.e.,} $\lambda_i=0, \forall i.$ However, as we show in Section \ref{bq_no_lambda} of the Appendix, the \BQ policy actually achieves the optimal instance-independent $O(\sqrt{T})$ regret bound for both full-information and bandit feedback setting when $\bm{\lambda} = \bm{0}$. 

As an immediate corollary of Theorem \ref{q_bd}, the following result shows that the under the action of the \textsc{BanditQ} policy, the target reward accrual rates are met asymptotically for each arm $i \in \mathcal{P}:$
%while incurring a reward violation penalty of $O(T^{3/4}).$ 
%Our result improves upon the $O(T^{5/6})$ violation penalty established by \citet[Theorem 1]{cai2018online} under independence assumptions.
\begin{proposition} \label{rate-prop}
	Upon setting $V=\sqrt{T},$ for any interval $\mathcal{I} \subseteq [T]$ such that $T^{3/4}=o(|\mathcal{I}|),$ the \textsc{BanditQ} policy in the full-information setting yields: 
	\[\liminf_{|\mathcal{I}| \to \infty} |\mathcal{I}|^{-1}\mathbb{E}\sum_{t \in \mathcal{I}} r_i(t)x_i(t) 
	  \geq \lambda_i, ~ \forall i \in \mathcal{P}.\]
\end{proposition}
See Appendix \ref{rate-prop-proof} for the proof. 
Although Theorem \ref{q_bd} gives an $O(T^{\nicefrac{3}{4}})$ regret bound compared to the minimax regret bound of $O(\sqrt{T \log N}),$ \citep[Theorem 4]{zhao2019stochastic} for the unconstrained problem, the next result shows that the proposed \BQ policy actually admits a substantially stronger $O(\sqrt{T})$ bound for the
\emph{average} regret, where the regret is averaged over the entire time horizon $T$. 
 
\begin{proposition} \label{avg-regret}
	In the full information setting, under the \BQ policy with $V=\sqrt{T},$ we have 	%\begin{eqnarray*}
		$\frac{1}{T}\sum_{t=1}^T \textrm{Regret}_t(\bm{x}^*) = O(\sqrt{NT}),$ for any $\bm{x}^* \in \Omega$ and $\mathbb{V}(T)=O(T^{\nicefrac{3}{4}}).$ 
	%\end{eqnarray*}
\end{proposition}
%See Appendix \ref{avg-regret-proof} for the proof. 
\begin{proof}
	Define $S_t^2 \equiv \sum_i \mathbb{E} Q_i^2(t).$ From Eq.\ \eqref{main_ineq}, for all $t \in [T],$  we have 
\begin{eqnarray*}
	S_t^2 + 2V \textrm{Regret}_t(\bm{x}^*) &\leq& 2t + 4 \sqrt{2 \sum_{\tau=1}^t S_\tau^2} + 4V \sqrt{2Nt} \\
	&\leq& 2T+ 4 \sqrt{2 \sum_{\tau=1}^T S_\tau^2}+4V \sqrt{2NT}.
\end{eqnarray*}
Summing up the above inequalites from $t=1$ to $t=T$ and defining $z_T\equiv \sqrt{\sum_{\tau=1}^T S_\tau^2},$   we obtain
\begin{eqnarray} \label{avg-ineq2}
 z_T^2 - 4Tz_T + 2V \sum_{t=1}^T\textrm{Regret}_t(\bm{x}^*) \leq 2T^2 + 4V\sqrt{2N}T^{3/2}.
\end{eqnarray}
Upon completing the square, we have  $z_T^2 - 4Tz_T = (z_T-2T)^2 -4T^2 \geq -4T^2. $ Hence, from \eqref{avg-ineq2}, we conclude that: 
\begin{eqnarray*}
	 \frac{1}{T}\sum_{t=1}^T\textrm{Regret}_t(\bm{x}^*) \leq 3\frac{T}{V} + 2\sqrt{2NT}.
\end{eqnarray*}
The final result follows upon setting $V=\sqrt{T}.$ 
\end{proof}
%The reader should compare the above bound with the  $\Omega(\sqrt{T\log N})$ minimax regret lower bound for stochastic rewards with the full-information feedback \citep[Theorem 4]{zhao2019stochastic}.  
 Finally, if one is only interested in achieving the target rate vector $\vec{\bm{\lambda}}$ while completely disregarding the regret, the following Proposition shows that the queue-length bound, and hence, the rate violation penalty given in Proposition \ref{q_bd} can be further improved to $O(\sqrt{T})$ upon setting $V=0.$ 
 \begin{proposition} \label{improved_bd}
 Setting $V=0,$ the cumulative constraint violation under the \BQ policy in the full-information setting can be bounded as follows:
 %the second and first moments of the state variables $\{\bm{Q}(t)\}_{t \geq 1}$ can be bounded as:
% \[\mathbb{E}Q_i^2(t) \leq 64Nt \implies \mathbb{E}Q_i(t) \leq 8\sqrt{Nt}, ~ \forall i \in \mathcal{P}, \forall t\geq 1.\]
\[ \mathbb{V}(T) \leq \max_i \mathbb{E}Q_i(T) \leq 6 \sqrt{T}. \]
 \end{proposition}
 \begin{proof}
 From Eq.\ \eqref{main-ineq3}, we have for any fixed $t$ and any $1\leq \tau \leq t:$
\begin{eqnarray} \label{new-eq}
	\sum_i \mathbb{E}Q_i^2(\tau) \leq 2t + 4 \sqrt{2\sum_{\tau=1}^t\sum_{i} \mathbb{E}Q_i^2(\tau) }~~ \forall t \geq 1, \forall i.
\end{eqnarray}
Summing up the above inequalities for $1\leq \tau \leq t$ and defining $z_t^2 \equiv  \sum_{\tau=1}^t \sum_{i} \mathbb{E}Q_i^2(\tau),$ we have 
\begin{eqnarray*}
	z_t^2 \leq 2t^2 + 4 \sqrt{2}tz_t.
\end{eqnarray*}
Solving the above quadratic inequality, we conclude that 
\begin{eqnarray*}
	\sqrt{\sum_{\tau=1}^t \sum_{i} \mathbb{E}Q_i^2(\tau)} = z_t \leq  6t. 
\end{eqnarray*}
Substituting the above bound in \eqref{new-eq} and using Jensen's inequality, we conclude that $\mathbb{E}Q_i(t) \leq 6\sqrt{t}, \forall i \in [N].$
\end{proof}
% See Appendix \ref{improved_bd_proof} for the proof. 
 %In the setting of Proposition \ref{improved_bd}, we obtain $\mathbb{V}(T) = O(\sqrt{NT})$.
  
  \paragraph{Sharper regret bound under a monotonicity assumption:} \label{stronger-bd}
  The regret and constraint violation bounds derived above hold unconditionally. We now show that the \BQ policy achieves the minimax optimal $O(\sqrt{T})$ regret under a mild monotonicity assumption on the queue length sequence stated below.  
  %These bounds can be strengthened under the following assumption.
%  \begin{assumption}[(Non-negativity of the regret)] \label{non-neg-regret}
%  	Assume that there exists some feasible distribution $\bm{p}^* \in \Omega(\bm{\vec{\lambda}})$ such that under the action of the \BQ policy, we have  $\textrm{Regret}_t(\bm{p}^*) \geq 0, \forall t \geq 1.$ Note that $\bm{p}^*$ need not be known to the policy.
%  \end{assumption}
\begin{assumption}[Monotonicity in expectation] \label{mon-q}
	Under the action of the chosen OLO subroutine, the sequence of variables $Q^2(t)\equiv \sum_i \mathbb{E}Q_i^2(t), t \geq 1$ are non-decreasing in $t$.
\end{assumption}
  %\citet{?} made a similar assumption.
  \begin{theorem} \label{mon-q-thm}
  	Under Assumption \ref{mon-q}, the regret of the \BQ policy in the full-information setting is bounded as \[\textrm{Regret}_t\leq \frac{5t}{V}+ 2 \sqrt{2Nt}, ~1\leq t \leq T.\]
  	In particular, with $V=\sqrt{T},$ we have $\textrm{Regret}_t = O(\sqrt{Nt})$ for any $t \in [T].$ 
  	\end{theorem}
%  	\begin{eqnarray*}
%		\textrm{Regret}_T(\bm{x}^*) = O(\sqrt{NT}), ~ \mathbb{V}(T) = O(N^{\nicefrac{1}{4}}\sqrt{T}), \forall \bm{x}^*\in \Omega.
%	\end{eqnarray*}
%  \end{theorem}
   See Appendix \ref{mon-q-thm-pf} for the proof. Assumption \ref{mon-q} is related to a stochastic monotonicity assumption. Many closely related Markov chains, \emph{e.g.,} the birth-death chain, which is a continuous-time model of a queue, with the zero initial states are known to be stochastically monotone \citep[Proposition 9.2.4]{ross1995stochastic} \citep[Theorem 6.1]{van1980stochastic}, \citep{keilson1977monotone}.  
   %assumption (Assumption \ref{q-mon-bandit2}), which we discuss further in the bandit setup.

 %All of the above result holds without imposing any additional restriction on the problem. The following proposition shows that if there exists an optimal action $\bm{x}^* \in \Omega$ such that the pseudo-regret $\textrm{Regret}_t(\bm{x}^*)$ is non-negative throughout (except, possibly, a finite number of rounds), then the regret can be improved to the optimal minimax rate $O(\sqrt{T}).$
 
 %\cmt{Discuss what happens when there exists an $\bm{x}^*$ with positive regret throughout.}

\section{\BQ policy with bandit feedback} \label{bandit-feedback}
%The additional difficulty in the bandit feedback setting compared to the full-information set-up is that if a policy selects an arm $I_t \in [N],$ on round $t,$ then 
Under bandit feedback, only the reward of the selected arm, \emph{i.e.,} $r_{I_t}(t),$ is revealed to the policy at the end of round $t$. The reader should compare this with the full-information setup where the entire reward vector $\bm{r}(t)$ is revealed irrespective of the action. To deal with the resulting in the \emph{exploration-vs-exploitation} trade-off in the limited information setup, we replace the full-information \textsc{OGA} policy \eqref{oga-update} with an adversarial MAB policy, proposed recently by \citet{putta2022scale}, that enjoys a \emph{scale-free} second-order regret bound similar to Eq.\ \eqref{data-dep-bd}. Their \emph{Follow-the-regularized-leader} (FTRL)-based MAB policy uses the standard inverse propensity score to estimate the reward vectors and employs a log-barrier regularizer in the FTRL algorithm with a carefully chosen learning rate schedule. The bandit arms are finally selected by mixing a uniform exploration component with the distribution obtained from the FTRL algorithm. For completeness, we describe the \BQ policy in the bandit information setting in Appendix \ref{BQ_bandit}. \citet{putta2022scale} showed that their proposed MAB policy works for \emph{any} real loss vector (unlike \emph{e.g.,} EXP3, which requires non-negative losses) and enjoys the following scale-free adaptive regret bound.  

\begin{theorem}[\cite{putta2022scale}] \label{ref_th2}
MAB Algorithm 1 of \cite{putta2022scale}, when run with the  oblivious linear reward sequence with coefficient vectors $\{\bm{g}_t\}_{t=1}^T,$ enjoys the following scale-free regret bound: 
\begin{eqnarray} \label{bandit_reg-bd}
	\textrm{Regret}_T = \tilde{O}\bigg(\sqrt{N\sum_{t=1}^T||\bm{g}_t||_2^2} + \max_{t \in [T]} ||\bm{g}_t||_\infty\sqrt{NT}\bigg).
\end{eqnarray}
\end{theorem}
It can be seen that the only essential difference between the above expression and that of the \textsc{OGA} regret bound in Eq.\ \eqref{data-dep-bd} is the presence of the additional term $\tilde{O}(\max_{t \in [T]} ||\bm{g}_t||_\infty\sqrt{NT})$ in the former. With a more careful analysis using martingales, our previous arguments go through with minimal changes. We now outline the main differences between the full information and the bandit setup. 

\textbf{Notation:} Let us encode the index of the selected arm $I_t$ on round $t$ by the one-hot encoded vector $\bm{X}(t)=\big(X_1(t), X_2(t), \ldots, X_N(t)\big) \in \{0,1\}^N$, where $X_i(t)= \mathds{1}(I_t=i), \forall i.$ Thus if $x_i(t)$ denotes the conditional probability that the $i$\textsuperscript{th} arm is pulled, we have $\mathbb{P}(X_i(t)=1|\mathcal{F}_{t-1})= 1-\mathbb{P}(X_i(t)=0|\mathcal{F}_{t-1})=x_i(t)$ and $\mathbb{E}(X_i(t)|\mathcal{F}_{t-1})=x_i(t), \forall i,t.$ 
%Hence the marginal distributions satisfy the relation $\mathbb{E}(X_i(t)|\mathcal{F}_{t-1})=x_i(t), \forall i,t.$

\paragraph{Queueing recursion and the auxiliary MAB problem:} Note that the queueing recursion \eqref{q-ev} for the full-feedback setting does not work in the case of Bandit feedback because the rewards of the unobserved arms are not revealed. However, it is straightforward to modify the recursion \eqref{q-ev} by replacing the sampling probabilities $\bm{x}(t)$ with the corresponding random realizations $\bm{X}(t).$ Hence, in the bandit setting, the queueing evolution for the $i$\textsuperscript{th} arm reads:
\begin{eqnarray} \label{q-ev-2}
	Q_i(t)=\big(Q_i(t-1)+ \lambda_i - r_i(t)X_i(t)\big)^+, ~Q_i(0)=0.
\end{eqnarray}
Eq.\ \eqref{q-ev-2} is well-defined in the bandit feedback setting as $X_i(t)=0$ if $i \neq I_t.$ Hence, the recursion \eqref{q-ev-2} does not depend on the reward of any arm which was not played. Next, analogous to the full-information setting (Eq.\ \eqref{reward_def}), the \textsc{BanditQ} policy defines an instance of an adversarial MAB problem $\Xi^{\textsc{Bandit}}$ where the surrogate reward of the $i$\textsuperscript{th} arm on round $t$ is defined as: 
\begin{eqnarray} \label{reward_def2}
	r'_i(t) \equiv \big(Q_i(t-1) + V\big)r_i(t), ~\forall i \in [N].
\end{eqnarray} 
As before, the surrogate rewards are not bounded \emph{a priori} due to the presence of the queueing variables. 
%Hence, we use a black box MAB policy with a scale-free second-order regret bound proposed in \citet{putta2022scale}, which is an FTRL policy with a logarithmic regularizer with a carefully chosen adaptive learning rate schedule. 
%For completeness, we define the entire policy in Appendix \ref{}.  

%\begin{theorem}[\cite[Algo 1]{putta2022scale}]  
%	d
%\end{theorem}
 
\subsection{Analysis} 
As before, the components of the surrogate reward gradients are given by $\bm{g}_{t,i}= r_i'(t)= \big(Q_i(t-1) + V\big)r_i(t).$
%Set $V=\Theta(\sqrt{T}), \forall t.$ 
%Next, we control the last term in the regret bound \eqref{bandit_reg-bd}. Note that $\max_t ||\bm{g}_t||_\infty = \max_{t=1}^T \max_i (Q_i(t-1)+V) \stackrel{(\textrm{a.s.})}{=} O(T),$ where we have used the fact that $Q_i(t) \stackrel{(\textrm{a.s.})}{\leq } T, \forall t\in [T].$ 
Using the quadratic potential function $\Phi(\cdot)$ defined in Eq.\ \eqref{potential_def}, and working identically up to step (c) of Eq.\ \eqref{main-ineq3}, we derive the following self-bounding inequality:
\begin{eqnarray} 
&&\sum_{i} \mathbb{E}Q_i^2(t) + 2V \textrm{Regret}_t (\bm{x}^*) \nonumber \\
	%&=&\sum_{i} \mathbb{E}Q_i^2(t) + 2V \sum_{\tau=1}^t  \mathbb{E}\sum_i r_i(\tau) \big(x_i^*-X_i(\tau)\big)\nonumber \\
	&\leq& 2t + 2 \mathbb{E}\big[ \textrm{Regret}^{\Xi^{\texttt{Bandit}}}_t\big]\nonumber \\
	&\stackrel{(a)}{\leq} & 2t + \tilde{O}\bigg(\sqrt{N\sum_{\tau=1}^t\sum_i \mathbb{E}Q_i^2(\tau) }+NV \sqrt{t} + \nonumber \\
	&&V\sqrt{Nt}+\sqrt{Nt}\mathbb{E}\big[\max_{i,\tau \in [t]}(Q_i(\tau))\big]\bigg)\label{main_bd2}\\
	&\stackrel{(b)}\leq&  2t + \tilde{O}\big(\sqrt{N\sum_{\tau=1}^t\sum_i \mathbb{E}Q_i^2(\tau) }+NV \sqrt{t} + \sqrt{N}t^{3/2}\big),\nonumber \\ \label{reg-bd9}
\end{eqnarray}
where, in step (a), we have used the regret bound from Theorem \ref{ref_th2}, and in step (b), we have used the trivial bound $Q_i(t) \leq t, \forall t \in [T], \forall i$. The following theorem gives the performance of the \BQ policy with bandit feedback.

\begin{theorem}\label{q_bd-bandit}
	%Upon setting $V=\Theta(\sqrt{T}), \forall t\geq 1$, 
	In the bandit feedback setting, the \BQ policy achieves the following regret and target rate violation bounds:
	\begin{eqnarray*}
			\textrm{Regret}_T &=& \tilde{O}(\max(\frac{T\sqrt{N}}{\sqrt{V}}, \frac{N^{\nicefrac{3}{4}}T^{\nicefrac{5}{4}}}{V},N\sqrt{T}))., \\
		\mathbb{V}(T) &=& \tilde{O}(\max(\sqrt{VT}, N^{\nicefrac{1}{4}}T^{\nicefrac{3}{4}})).
	\end{eqnarray*}
	In particular, upon setting $V=\sqrt{T},$ we obtain 
		\begin{eqnarray*}
		\textrm{Regret}_T = O(N^{\nicefrac{3}{4}}T^{\nicefrac{3}{4}}), ~ \mathbb{V}(T) = \tilde{O}(N^{\nicefrac{1}{4}}T^{\nicefrac{3}{4}}).
	\end{eqnarray*}
\end{theorem}
Compared to the full-information setting, the proof in the bandit setting uses a more sophisticated Martingale-based argument to control the maximum of the queueing process for bounding the second term in the regret expression \eqref{bandit_reg-bd}. 
%\begin{proof}
%See Appendix \ref{dia-bd} for the proof. 
%\section{Proof of Theorem \ref{q_bd-bandit}}
%\label{dia-bd}
To simplify the exposition, the proof of Theorem \ref{q_bd-bandit} is broken into three interrelated propositions.
We begin our analysis by first deriving a sublinear bound for $\mathbb{E}Q_i^2(t).$
\begin{proposition} \label{th5-pf1}
	Under the action of the \BQ policy with bandit feedback, we have 
	\[ \mathbb{E}Q_i^2(t)=\tilde{O}(\max(Vt, \sqrt{N}t^{\nicefrac{3}{2}})) , \forall i, t.\]
	Hence, using Jensen's inequality, we have $\mathbb{V}(T) =  \tilde{O}(\max(\sqrt{VT}, N^{\nicefrac{1}{4}}T^{\nicefrac{3}{4}})).$

\end{proposition}
\begin{proof}
	Recall that from Eqn.\ \eqref{reg-bd9} we have: 
	\begin{eqnarray*}
		&&\sum_{i} \mathbb{E}Q_i^2(t) + 2V \textrm{Regret}_t (\bm{x}^*) \leq 2t + \\
		&&\tilde{O}\big(\sqrt{N\sum_{\tau=1}^t\sum_i \mathbb{E}Q_i^2(\tau) }+NV \sqrt{t} + \sqrt{N}t^{3/2}\big).
	\end{eqnarray*}
	Using the fact that $r_i(t) \leq 1, \forall i,t,$ we have $\textrm{Regret}_t(x^*) \geq -t.$ Hence, from the above, we obtain
	\begin{eqnarray}\label{q-bd-bandit}
		&&\sum_{i} \mathbb{E}Q_i^2(t) \leq 2(V+1)t+ \nonumber \\
		&&\tilde{O}\big(\sqrt{N\sum_{\tau=1}^t\sum_i \mathbb{E}Q_i^2(\tau) }+NV \sqrt{t} + \sqrt{N}t^{3/2}\big),
	\end{eqnarray}
	which resembles Eqn.\ \eqref{ineq1} in the full-information setting. Defining $R(t) \equiv \sqrt{\sum_{\tau=1}^t \sum_{i=1}^N \mathbb{E}Q_i^2(\tau)},$ and working similarly as in the full-information setting, we have the following quadratic inequality:
	\begin{eqnarray} \label{R-bd-bandit}
		&&R^2(t) \leq 2(V+1)t^2 + \tilde{O}\big(\sqrt{N} tR(t)+NV t^{\nicefrac{3}{2}} + \sqrt{N}t^{\nicefrac{5}{2}}\big) \nonumber\\
		&&\emph{i.e.,} R(t) = \tilde{O}\big(\max(t\sqrt{V}, N^{\nicefrac{1}{4}}t^{\nicefrac{5}{4}})\big).
	\end{eqnarray}
	Substituting the above bound in \eqref{q-bd-bandit}, we conclude that for each $i \in [N]:$
	\begin{eqnarray*}
		\mathbb{E}Q_i^2(t) = \tilde{O}(\max(Vt, \sqrt{N}t^{\nicefrac{3}{2}})).
	\end{eqnarray*}
\end{proof}

The next proposition establishes a sublinear bound to the diameter $\mathbb{E}\big[\max_{i, t \in [T]}Q_{i}(t)\big]$, which appears on the RHS of \eqref{main_bd2}.
%of the queueing processes $\{\textbf{Q}(t)\}_{t=1}^T$. This result will be used to derive a sublinear regret bound for the \BQ policy. 

\begin{proposition}\label{uniform_bd_lemma}
Under the action of the \BQ policy, for any round $T\geq 1,$ we have the following bound for the expected maximum of the queueing processes
	\[ \mathbb{E}\big[\max_{i, t \in [T]}Q_{i}(t)\big] = \tilde{O}(\max(\sqrt{VT}, N^{\nicefrac{1}{4}}T^{3/4})).\]
\end{proposition}
The proof of Proposition \ref{uniform_bd_lemma} is technical and is given in Section \ref{unif-bd-pf} in the Appendix.
	Combining the above two results, the following proposition gives the worst-case regret bound for the \BQ policy under the bandit feedback.
	\begin{proposition} \label{reg-bd-bandit-2}
		The worst-case regret of the \BQ policy under the  bandit feedback is bounded as
		\begin{eqnarray*}
				\textrm{Regret}_T = \tilde{O}(\max(\frac{T\sqrt{N}}{\sqrt{V}}, \frac{N^{\nicefrac{3}{4}}T^{\nicefrac{5}{4}}}{V},N\sqrt{T})).
		\end{eqnarray*}
	\end{proposition}
	\begin{proof}
		From Eqn.\ \eqref{main_bd2}, we have 
		\begin{eqnarray} \label{main-bd43}
			\sum_i \mathbb{E}Q_i^2(T)+2V \textrm{Regret}_T(\bm{x}^*) \leq 2T  + \tilde{O}\bigg(\sqrt{N}R(T)+ \nonumber \\ NV \sqrt{T} + V\sqrt{NT}+\sqrt{NT}\mathbb{E}\big[\max_{i,\tau \in [T]}(Q_i(\tau))\big]\bigg),
		\end{eqnarray}
		where $R(T) \equiv \sqrt{\sum_{\tau=1}^T \sum_{i=1}^N \mathbb{E}Q_i^2(\tau)}$. Plugging in the upper bound for $R(T)$ from Eqn.\ \eqref{R-bd-bandit} and the diameter of the queueing process from Proposition \ref{uniform_bd_lemma}, we obtain:
		 \begin{eqnarray*}
		 	2V \textrm{Regret}_T(\bm{x}^*) = \tilde{O}(\max(T\sqrt{NV}, NV\sqrt{T}, N^{\nicefrac{3}{4}}T^{\nicefrac{5}{4}})).
		 \end{eqnarray*}
		 Hence,
		 \begin{eqnarray*}
		 	\textrm{Regret}_T(\bm{x}^*) = \tilde{O}(\max(\frac{T\sqrt{N}}{\sqrt{V}}, \frac{N^{\nicefrac{3}{4}}T^{\nicefrac{5}{4}}}{V},N\sqrt{T})).
		 \end{eqnarray*}

\end{proof}
Proposition \ref{th5-pf1} and Proposition \ref{reg-bd-bandit-2} taken together establish Theorem \ref{q_bd-bandit}. 

Following exactly the same arguments, the result in Proposition \ref{rate-prop} can be shown to hold in the bandit feedback setting as well. Finally, as in the full-information setting, we now discuss the case when one is only interested in satisfying the target rate constraints while disregarding the accrued rewards. The following proposition gives a bound on the cumulative violation in the bandit setting.

\begin{proposition} \label{rate-violation-bandit-no-reward}
	Setting $V=0,$ the cumulative constraint violation under the \BQ policy in the bandit setting can be bounded for any $T \geq 1$ as follows:
	\[ \mathbb{V}(T) \leq \max_i \mathbb{E}Q_i(T) = \tilde{O}(N^{\nicefrac{3}{8}}T^{\nicefrac{5}{8}}) . \]
\end{proposition}
%It can be seen that 
The above bound is slightly worse compared to the $O(\sqrt{T})$ bound in the full-information setting (Proposition \ref{improved_bd}). 
See Section \ref{rate-violation-bandit-no-reward-proof} in the Appendix for the proof of Proposition \ref{rate-violation-bandit-no-reward}.

\begin{figure*}[t]
  \centering
  \begin{minipage}[b]{0.3\linewidth}
   \centering
    \includegraphics[width=\linewidth]{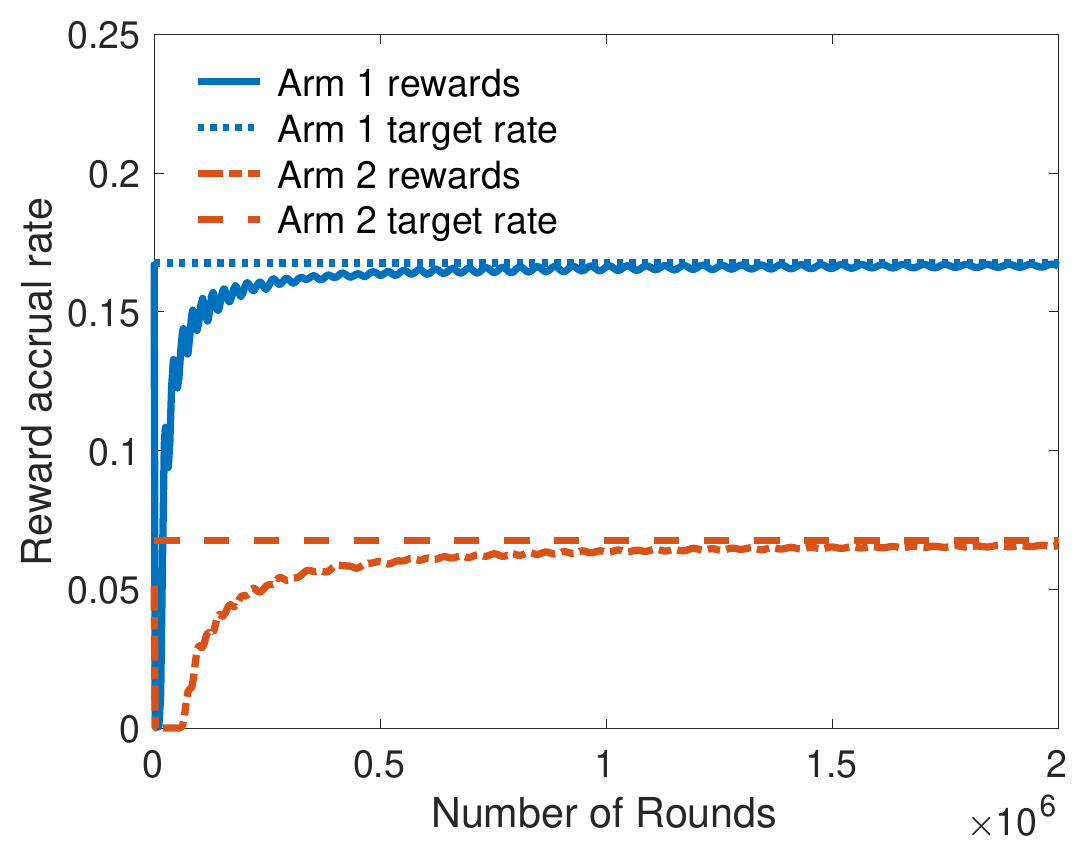}
   \caption{\small{Reward accrual rates in the full-information setting}}
   \label{rew_full}
  \end{minipage}
   \begin{minipage}[b]{0.3\linewidth}
   \centering
    \includegraphics[width=\linewidth]{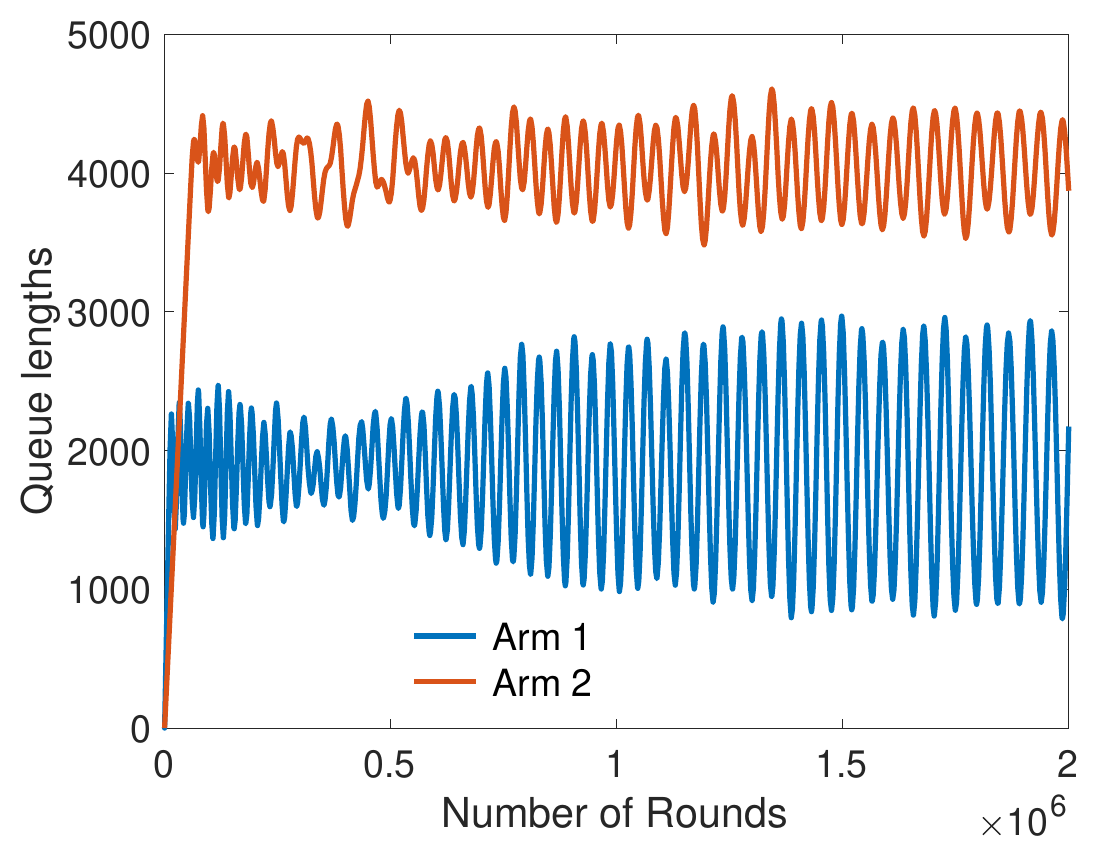}
   \caption{\small{Queue lengths in the full-information setting}}
   \label{q_full}
  \end{minipage}
   \begin{minipage}[b]{0.3\linewidth}
   \centering
    \includegraphics[width=\linewidth]{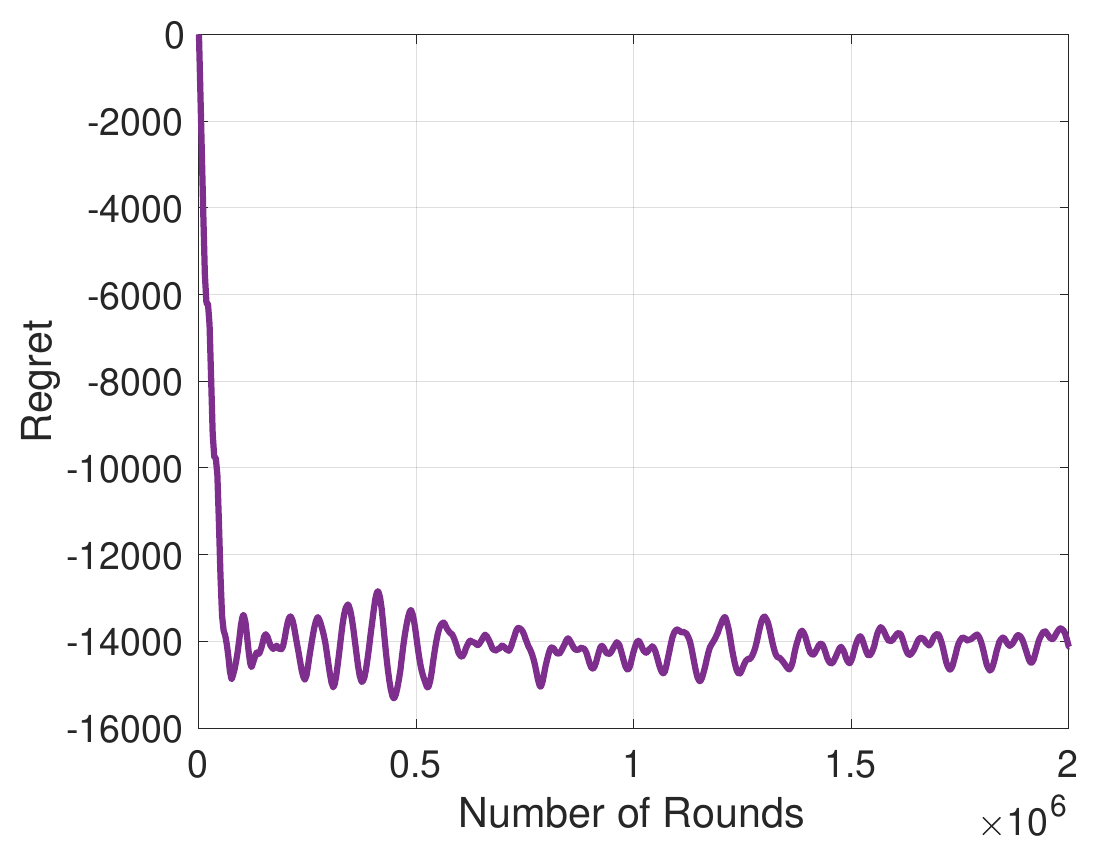}
   \caption{\small{Regret of \BQ in the full-information setting}}
   \label{reg_full}
  \end{minipage}
  \begin{minipage}[b]{0.3\linewidth}
   \centering
    \includegraphics[width=\linewidth]{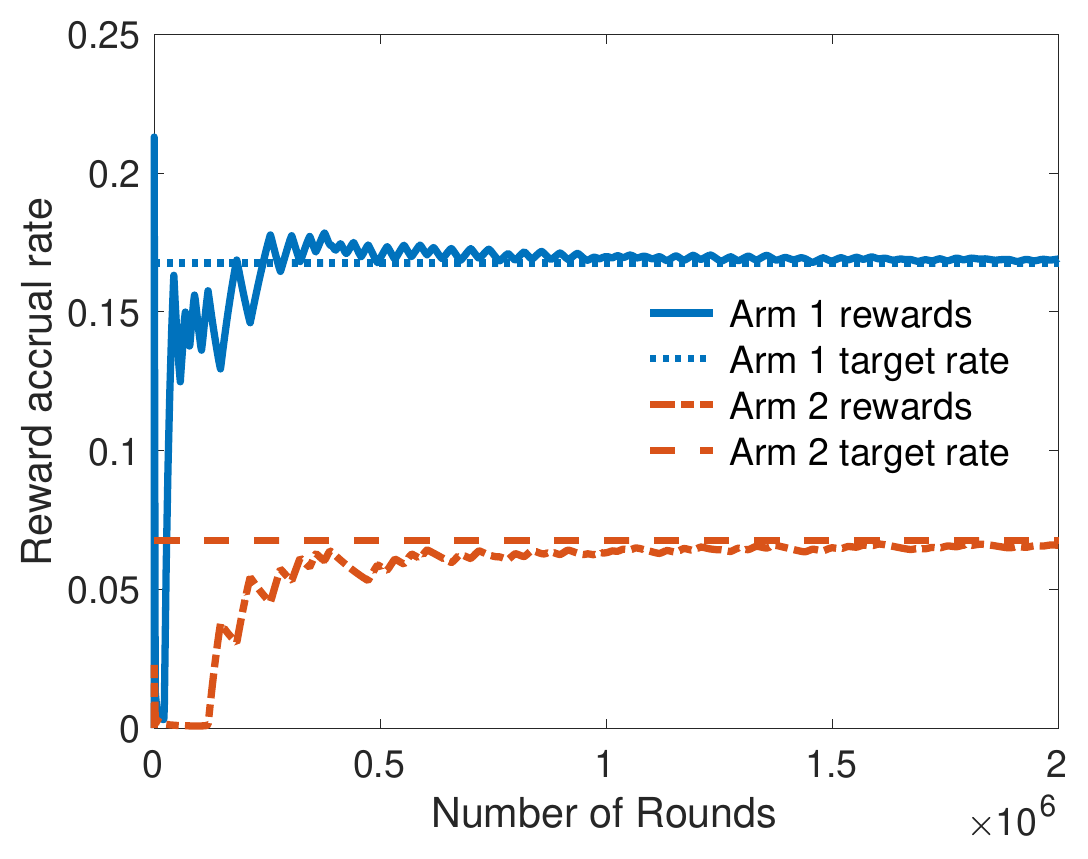}
   \caption{\small{Reward accrual rates in the bandit feedback}}
   \label{rew_bf}
  \end{minipage}
  %\hfill
  \begin{minipage}[b]{0.3\linewidth}
   \centering
    \includegraphics[width=\linewidth]{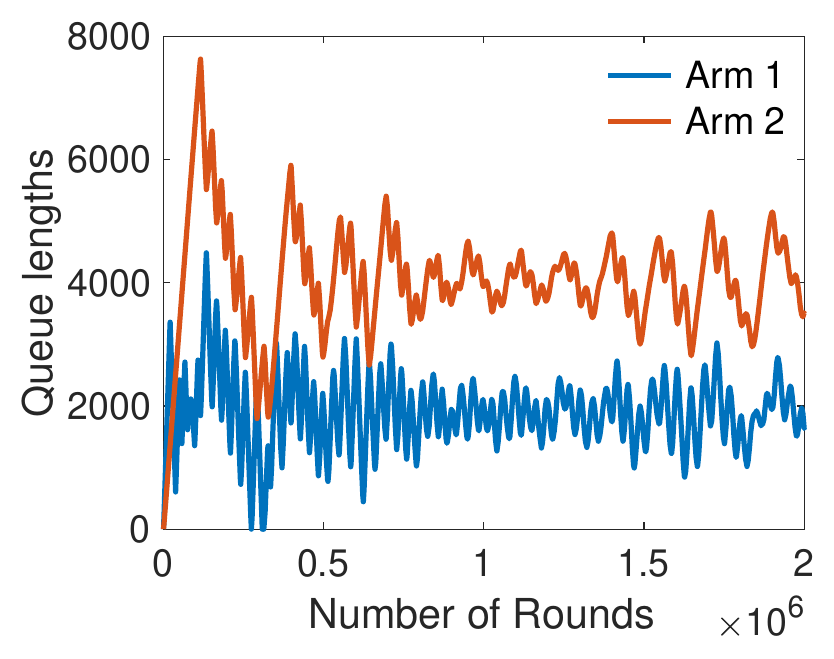}
   \caption{\small{Queue lengths in the bandit feedback setting}}
   \label{q_bf}
  \end{minipage}
 %\hfill
   \begin{minipage}[b]{0.3\linewidth}
   \centering
    \includegraphics[width=\linewidth]{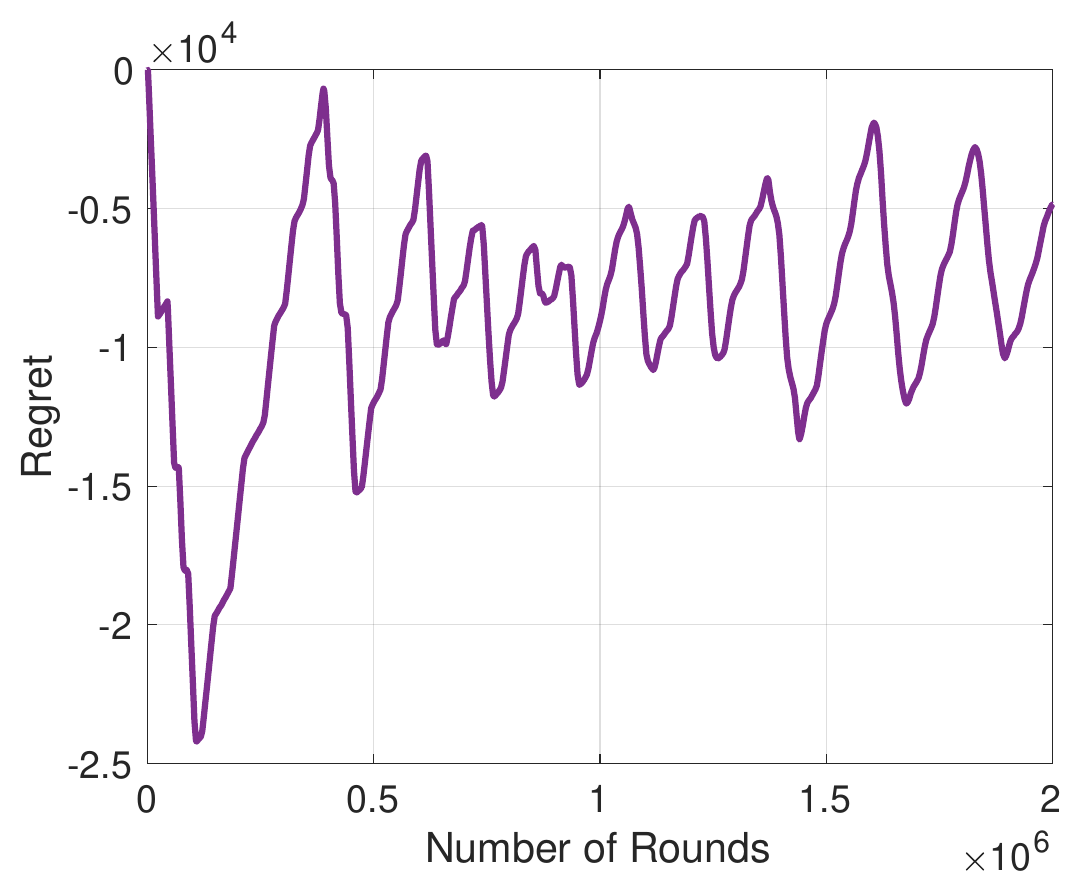}
   \caption{\small{Regret of \BQ in the bandit feedback setting}}
   \label{reg_bf}
  \end{minipage}
\end{figure*}

\paragraph{Remarks:} 
\iffalse 1. When all target rates are zero (\emph{i.e.,} $\vec{\bm{\lambda}}=\vec{\bm{0}}$), the fair prediction problem reduces to the classic MAB problem, which is known to have a minimax regret bound of $O(\sqrt{NT})$ \citep{lattimore2020bandit}. Hence, improving the regret bound given by Theorem \ref{q_bd-bandit} might be possible. The main challenge for proving an $O(\sqrt{T})$ regret bound appears to be controlling the term $\mathbb{E}(\max_{i,t}Q_i(t))$ in the regret expression \eqref{main_bd2}.  
\fi
%We leave the question of the tightness of the above regret bound as an interesting open problem. 
  Technically, the scale-free regret bound given in Theorem \ref{ref_th2} was derived for \emph{oblivious} adversaries, which fixes the entire sequence of reward vectors at $t=0$. However, in our case, the surrogate reward vector $\bm{r}'(t)$ in Eqn.\ \eqref{reward_def2} is determined by the past actions of the policy through the variable $\bm{Q}(t)$. To see why we can still use the regret bound \eqref{bandit_reg-bd}, note that the surrogate reward $\bm{r}'(t)$ does not depend on the current action $\bm{X}(t).$ Hence, we can invoke the regret bound for an imaginary adversary that decides the reward vector $\bm{r}'(t)$ at the end of round $t-1$. Since the reward on round $t$ does not affect the previous actions of the policy, the regret bound \eqref{bandit_reg-bd} applies to our problem.

\section{Experiments} \label{simulation}
%In this section, we numerically compare the performance of \BQ policy with a set of state-of-the-art benchmarks \cmt{select the benchmarks}. 
\paragraph{Simulation Setup:}
We consider a problem instance with $N=5$ arms and $k=2$ protected classes consisting of the first and the second arm. We arbitrarily set the mean reward vector of the arms to $\bm{\mu} = \begin{pmatrix} 0.335, 0.203, 0.241, 0.781, 0.617 \end{pmatrix},$ and the target reward rates for the first and the second arm to $\lambda_1= 0.167$ and $\lambda_2= 0.067$ respectively. From Eq.\ \eqref{feas-constr}, it can be verified that the required rates are feasible for this problem. Clearly, Arm \# $4$ is the most rewarding among the five arms. We simulate the \BQ policy for $T=2\times 10^6$ rounds upon setting the parameter $V =\sqrt{T} \approx 1414.$ We write a custom optimizer, described in Appendix \ref{opt_implementation}, for efficiently implementing the optimization subroutines. The simulation code has been made publicly available \citep{BanditQ_code}.
% for the \BQ policy in the bandit feedback set up. 
%We now report the performance of the \BQ policy under both full-information and bandit feedback. 
%\cmt{Compare the performance with an oracle policy that knows the rewards and hence, knows the optimal fraction of pulls.} 
\paragraph{Discussion:}
Figures \ref{rew_full}, \ref{q_full}, and \ref{reg_full} show the performance of the \BQ policy in the full-information set-up. Figure \ref{rew_full} shows that the protected arms, Arm 1 and Arm 2, asymptotically meet their target rates. Observe that since both Arm 1 and Arm 2 have sub-optimal expected rewards, they would have received asymptotically zero reward rates under the action of an unfair prediction policy, such as UCB. Figure \ref{q_full} shows the evolution of the queue length variables, and Figure \ref{reg_full} shows the regret of the \BQ policy in the full-information setting. Negative values of the regret suggest that the cumulative reward of the \BQ policy exceeds the reward achieved by the static benchmark policy, which is forced to take actions from the restricted set $\Omega$ on \emph{all} rounds - a constraint that the \BQ policy does not need to respect on every rounds. Figures \ref{rew_bf}, \ref{q_bf}, and \ref{reg_bf} show the corresponding plots in the bandit feedback setting. As expected, in the case of bandit feedback, the variables exhibit greater empirical variance compared to their full-information counterpart due to the limited availability of information. However, the \BQ policy achieves the target rates in this case as well. See Section \ref{addl_sim2} in the Appendix for a similar experiment with $N=1000$ arms.      
%In both cases, we find that the observed regret of the \BQ policy is much better than the theoretical bounds. This observation can be justified by the fact that we, in fact, used adversarial MAB policies in the benign i.i.d. stochastic setting which can take actions from a larger set $\Delta_N$. 
%The regret plots suggest that it might be possible to prove sublinear regret bound for the \BQ policy in the case of adversarial rewards as well. 
\paragraph{Additional experiments:} A comparison of the \BQ policy with a UCB-based oracle policy, proposed by \citet{li2019combinatorial} has been given in Appendix \ref{addl_sim1}. The oracle is assumed to know a feasible fraction of pulls to achieve the target rates. The plot in Figure \ref{rew-comp} shows that the proposed \BQ policy achieves more cumulative rewards compared to the oracle policy as it decides its actions adaptively.

\section{Conclusion and open problems}   \label{open_problem}
%It is currently open whether the rate region can be extended beyond Eq.\ \eqref{nec-suff} in the adversarial setting. 
In this paper, we proposed a black-box reduction from the fair bandits problem to the unconstrained bandit problem and upper bounded the regret and cumulative target rate violations. 
%As a consequence, any improvement in the regret bound for MAB 
Since we use adversarial MAB policies as subroutines, it is reasonable to conjecture that the proposed \textsc{BanditQ} policy would work well in the adversarial setting as well. Substantiating this statement would be an interesting research direction \citep{sinha2023playing}.  
Improving the regret and rate violation bounds by, \emph{e.g.,} working with a different Lyapunov function would be practically useful.
Finally, coming up with sharper instance-dependent regret bounds would be interesting as well.
%Extending the \textsc{BanditQ} policy to the bandit information setup would also be of substantial interest. 
%Finally, designing an anytime version of the policy that does not need to know the  horizon length $T$ in advance would be practically useful.
\section{Acknowledgement}
The work was supported in part by a Google India faculty Research award and in part by a US-India NSF-DST collaborative grant coordinated by IDEAS-Technology Innovation Hub (TIH) at the Indian Statistical Institute, Kolkata. The author gratefully acknowledges a lively discussion with Prof. Anurag Kumar from IISc while formulating the problem.
\clearpage
\bibliography{OCO}
\bibliographystyle{unsrtnat}
\clearpage
%\begin{center}
%\centering
%\textbf{\large{Supplementary material for "\texttt{BanditQ:} Fair Multi-Armed Bandits with Guaranteed Rewards per Arm" }}
%\end{center}
\newpage
\onecolumn
\title{ Appendix for \\
\textsc{BanditQ:} Fair Bandits with Guaranteed Rewards}
\maketitle
\appendix
%\vspace{-140pt}
%h
%\vspace{-85pt}
%\section{Appendix} \label{appendix}
\section{On the feasibility assumption}\label{feas-sec}
Throughout the paper, we assume that the target rate vector $\bm{\vec \lambda}$ is feasible. In practice, we can ensure the feasibility by estimating the expected rewards from past data and requiring that condition \eqref{feas-constr} is strictly satisfied with a reasonable margin. To put it quantitatively, let $\hat{\bm{\mu}}$ be the estimated expected reward vector where it is known that $||\hat{\bm{\mu}} -\bm{\mu}||_\infty \leq \epsilon,$ for a small error bound $\epsilon \geq 0$. Then, for the required reward rate vector $\bm{\vec \lambda}$ to be feasible, using the first-order Taylor's series expansion, it is sufficient that: 
\begin{eqnarray} \label{feasibility_test}
	\sum_i \frac{\lambda_i}{\hat{\mu_i}} + \epsilon \sum_i \frac{\lambda_i}{\hat{\mu_i}^2} \leq 1.
\end{eqnarray}
Although the estimated mean rewards can reasonably be used for determining the feasibility of the required reward rates, they cannot possibly be used for the online selection of the arms with no regret, as even a small constant error in the estimated rewards may lead to a linear regret. 

\section{$O(\sqrt{T})$ Regret of the \BQ Policy with no target rates} \label{bq_no_lambda}
We now consider the classical and special case when there is no specific target rates for any of the arms, \emph{i.e.,} $\lambda_i=0, \forall i.$ Hence, from Eqn.\ \eqref{q-ev}, we have that $Q_i(t)=0, \forall i,t.$ Furthermore, with $\bm{\lambda}=\bm{0},$ the comparator class $\Omega$ coincides with the set of all probability distributions over $N$ arms ($\Delta_N$). We have the following result
\begin{proposition} \label{zero-target-rate}
	With no pre-specified target reward rates, \emph{i.e.,} $\bm{\lambda}=\bm{0},$ the \BQ policy achieves regret bounds of $O(\sqrt{Nt})$ and $\tilde{O}(N\sqrt{t})$ for the full-information and bandit feedback settings, respectively. 
\end{proposition}
Intuitively, the above result can be understood from the fact that, in this case, the surrogate rewards $\bm{r}'(t)$ of the \BQ policy is simply a scaled version of the original rewards $\bm{r}(t).$ See below for a formal proof.

\textbf{Proof:}
\paragraph{Full-information setting:}
From the regret decomposition inequality \eqref{main-ineq3}, we have that 
\begin{eqnarray*}
	2V \textrm{Regret}_t(x^*) \leq 2t + 4V \sqrt{2Nt}.
\end{eqnarray*}
Setting $V=\sqrt{T},$ we have that 
\begin{eqnarray*}
	\textrm{Regret}_t(x^*) \leq \frac{t}{V}+ 2 \sqrt{2Nt} = O(\sqrt{Nt}).
\end{eqnarray*}

\paragraph{Bandit information setting:} The proof is almost identical to the full-information case. Setting $Q_i(t)=0, \forall i,t,$ in  the regret decomposition inequality \eqref{main_bd2}, we have that 
\begin{eqnarray*}
	2V \textrm{Regret}_t(x^*) \leq 2t + \tilde{O}(NV\sqrt{t}+V\sqrt{Nt}). 
\end{eqnarray*}
Setting $V=\sqrt{T},$ the above yields 
\begin{eqnarray*}
	\textrm{Regret}_t(x^*) = \tilde{O}(N\sqrt{t}).
\end{eqnarray*}

\section{Proof of Proposition \ref{rate-prop}} \label{rate-prop-proof}
%\begin{proof}
	
Using Proposition \ref{q_bd}, we have that $\mathbb{E}Q_i(t) \stackrel{\textrm{(Jensen's ineq.)}}{\leq}\sqrt{\mathbb{E}Q_i^2(t)} = O(N^{1/4}T^{3/4}), ~\forall i\in \mathcal{P}, t\in [T].$ Let $\mathcal{I} \subseteq [1,T]$ be any sub-interval of length $l = |\mathcal{I}|$. 	
	Substituting the above bound in Eq.\ \eqref{q-len-bd}, we have for any $i \in \mathcal{P}:$
	\begin{eqnarray*}
	\inf_{\mathcal{I}}\mathbb{E}\big(\nicefrac{1}{|\mathcal{I}|}\sum_{z \in \mathcal{I}} r_i(z) x_i(z)\big) \geq \lambda_i  - O(\frac{N^{1/4}T^{3/4}}{l}),
	\end{eqnarray*}
	which gives a finite-time guarantee for the expected reward accrual rate for each arm in the protected set $\mathcal{P}.$
	Hence, as long as $\nicefrac{T^{3/4}}{l} \to 0,$ we have
	\begin{eqnarray*}
		\liminf_{|\mathcal{I}| \to \infty} |\mathcal{I}|^{-1}\mathbb{E} \big[\sum_{t \in \mathcal{I}} r_i(t)x_i(t)\big] \geq \lambda_i, ~\forall i \in \mathcal{P}.
	\end{eqnarray*} 

\section{Proof of Theorem \ref{mon-q-thm}} \label{mon-q-thm-pf}
Let $\bm{x}^*$ be an optimal fixed feasible randomized action. From Eqn.\ \eqref{main-ineq3}, we have that 
\begin{eqnarray*}
	2V \textrm{Regret}_t(\bm{x}^*) \leq  4\sqrt{2\sum_{\tau=1}^t\sum_i \mathbb{E}Q_i^2(\tau)}- \sum_{i} \mathbb{E}Q_i^2(t)+ 2t +4V\sqrt{2Nt}.
\end{eqnarray*}
Define $Q^2(\tau)= \sum_i \mathbb{E}Q_i^2(\tau), \forall \tau \geq 1.$ Using the monotonicity assumption \ref{mon-q}, the above inequality yields
\begin{eqnarray*}
	2V \textrm{Regret}_t(\bm{x}^*) &\leq& \underbrace{4\sqrt{2t}Q(t)- Q^2(t)}_{(A)}+ 2t +4V\sqrt{2Nt} \\
	&\stackrel{(a)}{\leq} & 10t + 4V\sqrt{2Nt}.
\end{eqnarray*}
where in (a), we have upper-bounded the quadratic (A), which is of the form $ax-x^2 $, by $a^2/4 \equiv 8t$. Hence, we have
\begin{eqnarray*}
	\textrm{Regret}_t \leq \frac{5t}{V}+ 2 \sqrt{2Nt}.
\end{eqnarray*}

%Upon completing the square, we conclude that: 
%\begin{eqnarray*}
%	 z_t \leq .
%\end{eqnarray*}
%The result follows upon setting $V=\Theta(\sqrt{T}).$ 
%\subsection{Bound on the diameter of the queueing processes $\{\textbf{Q}(t)\}_{t=1}^T$} 
\section{Proof of Proposition \ref{uniform_bd_lemma}} \label{unif-bd-pf}
\begin{proof}
Using Eq.\ \eqref{main_ineq} and the fact that $|r_i(t)|\leq 1, \forall i, t,$ we have the following sample-path wise bound on the square of the queue lengths:
	\begin{eqnarray} \label{sample-path}
		\sum_i Q_i^2(t) &\leq& 2(V+1)t + 2 \sum_{\tau=1}^t \sum_{i} Q_i(\tau-1) \big(\lambda_i-r_i(\tau) x_i^* \big)+2\textrm{Regret}^{\Xi}_t \nonumber\\
		&\leq& 2(V+1)t + \tilde{O}(\sqrt{N\sum_{\tau=1}^t\sum_i Q_i^2(\tau) }+NV \sqrt{t} + \sqrt{Nt}\max_{i, \tau \in [1,t]}Q_i(\tau)) + 2\sum_i M_t^i, \label{q-bd-bandit-ref} \\
		&\stackrel{(a)}{\leq}& 2(V+1)t + \tilde{O}(\sqrt{N\sum_{\tau=1}^t\sum_i Q_i^2(\tau) }+NV \sqrt{t} + \sqrt{N}t^{3/2}) + 2\sum_i M_t^i,
	\end{eqnarray}
	where in the above, we have substituted the upper-bound to the regret of the surrogate problem from Eq.\ \eqref{bandit_reg-bd}, (as in Eqn.\ \eqref{reg-bd9}), used the fact that $Q_(\tau) \leq \tau$, and for each $i \in [N],$ we have defined the stochastic process $\{M_t^i\}_{t \geq 1}$ as follows:
	\begin{eqnarray} \label{mg-def}
		M_t^i = \sum_{\tau=1}^t  Q_i(\tau-1) \big(\lambda_i'-r_i(\tau) x_i^* \big),~ t \geq 1.
	\end{eqnarray}
where $\lambda_i' \stackrel{(\textrm{def.})}{=} x_i^*\mathbb{E}r_i(\tau) = x_i^*\mu_i \geq \lambda_i.$
	%Hence, recalling that $V=\Theta(\sqrt{T}),$ 
	Taking the maximum of both sides with respect to all rounds $t \in [T]$ for some $T \geq 1,$ we have
	\begin{eqnarray*}
		\max_{i, t \in [T]}Q_i^2(t) \leq 2VT+ \tilde{O}(\sqrt{N\sum_{\tau=1}^T\sum_i Q_i^2(\tau) }+NV\sqrt{T}+ \sqrt{N}T^{3/2}) + 2\sum_i \max_{t \in [T]} M_t^i.
	\end{eqnarray*}
	Taking the expectation of both sides of the above inequality, we obtain
	\begin{eqnarray} \label{mg-bound}
		\mathbb{E}\big[\max_{i, t \in [T]}Q_i^2(t)\big] &\leq& 2VT+ \tilde{O}(\sqrt{N\sum_{\tau=1}^T\sum_i \mathbb{E}Q_i^2(\tau) }+ \sqrt{N}T^{3/2}) + 2\sum_i \mathbb{E}\big[\max_{t\in [T]} M_t^i \big] \nonumber \\
		&\stackrel{(a)}{\leq}& 2VT+ \tilde{O}\big(\max(T\sqrt{V}, N^{\nicefrac{1}{4}}T^{\nicefrac{5}{4}})\big)+ \tilde{O}(\sqrt{N}T^{3/2})+ 2\sum_i \mathbb{E}\big[\max_t M_t^i \big] \\
		&=& \tilde{O}(\max(VT, \sqrt{N}T^{3/2})) + 2\sum_i \mathbb{E}\big[\max_t M_t^i \big], 
	\end{eqnarray}
	where in step (a), we have used the bound for $R(T)$ from Eqn.\ \eqref{R-bd-bandit}.
	Next, we claim that each of the processes $\{M_t^i\}_{t\geq 1}$ is a zero-mean martingale process with respect to the natural filtration $\{\mathcal{F}_\tau\}_{\tau \geq 1}$. This follows from the definition \eqref{mg-def} as $Q_i(\tau-1) \in \mathcal{F}_{\tau-1}$ is pre-visible and the random variable $r_i(\tau)$ is independent of $\mathcal{F}_{\tau -1}$ s.t. $\mathbb{E}(\lambda_i' - r_i(\tau)x_i^*)=0.$  
	%Using classic results 
	Using the $L^2$ maximum inequality for Martingales \citep[Theorem 4.4.4]{durrett2019probability}, \citep[Theorem 3.4]{doob1953stochastic}, \citep{dubins1988sharp}, we have
	%we know that the diameter of a Martingale with a last term is bounded by twice the square root of the variance of the last term. Hence,
	\begin{eqnarray} \label{M-bd1}
		\mathbb{E}[\max_{t \in [T]} M^i_t] \leq 2\sqrt{\mathbb{E}(M^i_T)^2}.
	\end{eqnarray}
	Since $\{M_t\}_{t \geq 1}$ is a zero-mean martingale sequence, using the Pythagorean formula for martingales \cite[Eq. (b), Section 12.1]{williams1991probability} and the fact that $|\lambda_i' - r_i(\tau) x_i^*| \leq 1,$ we have
	\begin{eqnarray} \label{M-bd2}
		\mathbb{E}(M_T^i)^2 &\leq&  \sum_{\tau=1}^T \mathbb{E}Q_i^2(\tau-1) \\
		&\leq& R^2(T), \nonumber
	\end{eqnarray}
	where we have defined $R(T) \equiv \sqrt{\sum_{\tau=1}^T \sum_{i=1}^N \mathbb{E}Q_i^2(\tau)}$. Combining the above with Eq.\ \eqref{mg-bound}, we obtain the desired bound for the diameter of the queueing process:
	\begin{eqnarray*}
		\mathbb{E}\big[\max_{i, t \in [T]}Q_i^2(t)\big] \leq \tilde{O}(\max(VT, \sqrt{N}T^{3/2})) + O(R(T)) = \tilde{O}(\max(VT, \sqrt{N}T^{3/2})),
	\end{eqnarray*}
	where, we have again used the bound for $R(T)$ from Eqn.\ \eqref{R-bd-bandit}.
	The result stated in the lemma finally follows from an application of Jensen's inequality.
	\end{proof}

\section{Proof of Proposition \ref{rate-violation-bandit-no-reward}} \label{rate-violation-bandit-no-reward-proof}
Setting $V=0$ and plugging in the bound from Proposition \ref{uniform_bd_lemma}, we have the following bound from \eqref{main_bd2} for any round $1\leq t \leq T:$
\begin{eqnarray} \label{q-bd-bandit-pf}
	\sum_i \mathbb{E}Q_i^2(t) \leq 2T + \tilde{O}\big(\sqrt{N \sum_{t=1}^T \sum_i \mathbb{E}Q_i^2(t)}) + N^{\nicefrac{3}{4}}T^{\nicefrac{5}{4}}\big).
\end{eqnarray}
Define $z_T^2 \equiv \sum_{i}\sum_{t=1}^T \mathbb{E}Q_i^2(t).$ Summing up the inequalities \eqref{q-bd-bandit-pf} from $t=1$ to $t=T,$ we obtain 
\begin{eqnarray*}
	z_T^2 \leq 2T^2 + \tilde{O}(\sqrt{N}Tz_T + N^{\nicefrac{3}{4}}T^{\nicefrac{9}{4}}) \implies z_T = \tilde{O}(N^{\nicefrac{3}{8}}T^{\nicefrac{9}{8}}).
\end{eqnarray*}
Plugging in the above bound in \eqref{q-bd-bandit-pf}, we conclude that
\[\sum_i \mathbb{E}Q_i^2(T) = \tilde{O}(N^{\nicefrac{3}{4}}T^{\nicefrac{5}{4}}) \stackrel{\textrm{(Jensen's ineq.)}}{\implies} \mathbb{E}Q_i(T) = \tilde{O}(N^{\nicefrac{3}{8}}T^{\nicefrac{5}{8}}),~\forall i \in [N]. \]

\iffalse

\subsection{Proof of Theorem \ref{q-mon-bandit}} \label{q-mon-bandit-pf}

Using Eqn.\ \eqref{q-bd-bandit-ref}, taking maximum of both sides w.r.t. $t\in [T]$ and then taking expectation, we have 
\begin{eqnarray*}
	\mathbb{E}[\max_{i, t \in [T]} Q_i^2(t)] \leq 2(V+1)T + \tilde{O}(\sqrt{N\sum_{\tau=1}^T\sum_i \mathbb{E}Q_i^2(\tau) }+NV \sqrt{T} + \sqrt{NT}\mathbb{E}[\max_{i, t \in [1,T]}Q_i(t)]) + 2\sum_i \mathbb{E}[\max_{t \in [T]} M_t^i]
\end{eqnarray*}
Using Jensen's inequality on the LHS and the inequalities \eqref{M-bd1} and \eqref{M-bd2} to bound the right-most expectation in the above, we have the following quadratic inequality: 
\begin{eqnarray*}
	x^2 \leq \tilde{O}(\sqrt{NT}x+VT +NV\sqrt{T}+ \sqrt{N\sum_{t=1}^T \sum_i  \mathbb{E}Q_i^2(\tau)}),
\end{eqnarray*}
where we have defined $x \equiv \mathbb{E}[\max_{i, t \in [1,T]}Q_i(t)]$ and used Jensen's inequality for the concave square-root function on the right. Solving the above quadratic inequality, we have
\begin{eqnarray*}
	x \leq \tilde{O} (\sqrt{VT}).
\end{eqnarray*}
\edit{This seems to mess up the bound!}

\subsection{Proof of Theorem \ref{mon-bandit}} \label{mon-bandit-pf}
Since the sequence of r.v.s $\{Q_i(t)\}_{t \geq 1}$ are assumed to be stochastically monotone, there exists a probab 
\fi
\section{Pseudocode for the \BQ policy in the bandit feedback set up} \label{BQ_bandit}
As discussed in the main text, the \BQ policy in the Bandit feedback setting uses the scale-free MAB algorithm of \citet{putta2022scale} in conjunction with the surrogate reward function defined in Eq.\ \eqref{reward_def2}. The complete pseudocode of the \BQ policy is given below in Algorithm \ref{fair-MAB-bandit-info}. 
\begin{algorithm}
\caption{\BQ Policy in the Bandit-feedback setting}
\label{fair-MAB-bandit-info}
\begin{algorithmic}[1]
\State \algorithmicrequire{ Target reward rate vector $\bm{\vec{\lambda}}$, $\eta \gets N, \gamma \gets 1/2$, Regularizer $F(q)= \sum_{i=1}^N (f(q(i)-f(1/N)),$ where $f(x)=-\log(x).$} 
\State $\bm{Q} \gets \bm{0}, \bm{p} \gets [1/N, 1/N, \ldots, 1/N], V\gets \sqrt{T}, S\gets 1, \bm{\tilde{R}}\gets 0.$ \algorithmiccomment{\emph{Initialization}}
\ForEach {round $t=1:T$:}
\State $\bm{x} \gets (1-\gamma)\bm{p} + \gamma/N$. \algorithmiccomment{\emph{Updating the sampling distribution}}
\State Sample an arm $I_t \in [N]$ from the distribution $\bm{x}$.  
\State Observe the reward of the selected arm $r_{I_t}(t)$\algorithmiccomment{\emph{Bandit feedback}}
%\ForEach {arms $i \in \mathcal{P}$:}
\State 
%\begin{eqnarray*} 
	$Q_i=\big(Q_i+ \lambda_i - r_i(t)\mathds{1}(I_t=i)\big)^+, ~\forall i\in \mathcal{P}. $\algorithmiccomment{\emph{Updating the queues}}
%\end{eqnarray*}
%\EndForEach
\State $r'_i \gets \big(Q_i + V\big)r_i(t)\mathds{1}(I_t=i), ~\forall i$ \algorithmiccomment{\emph{Computing the surrogate rewards}}
\State $\tilde{r}_i \gets \frac{r'_i}{x_i} \mathds{1}(I_t=i)$ \algorithmiccomment{\emph{Estimating the rewards via the inverse propensity scores (IPS)}}
\State $\bm{\tilde{R}}\gets \bm{\tilde{R}} + \bm{\tilde{r}}$ \algorithmiccomment{\emph{Updating the cumulative estimated surrogate rewards}}
\State $\gamma \gets \min(1/2, \sqrt{N/t}).$ 
%\State $S \gets S + ||\bm{r}'(t)||^2.$ \algorithmiccomment{\emph{Accumulating the norm of past gradients}}
\State $S \gets S+ \eta^{-1}\sup_{q \in \Delta_N}(\langle \tilde{\bm{r}}, \bm{q}-\bm{p} \rangle -\textrm{Breg}_F(\bm{q}||\bm{p}).$  \label{opt1}
\State $\eta \gets N/S$ \algorithmiccomment{\emph{Adaptively choosing the learning rate}}
\State $\bm{p}\gets \arg \min_{\bm{q} \in \Delta_{N}} \big[ F(\bm{q}) - \eta \langle \bm{q}, \bm{\tilde{R}}\rangle \big]$ \algorithmiccomment{ \emph{The \texttt{FTRL} step}} \label{opt2}
	%\State $\bm{x}\gets \Pi_{\Delta_N}\bigg(\bm{x}+ \frac{\bm{r}'(t)}{\sqrt{2S}} \bigg)$ \algorithmiccomment{\emph{Implementing the online gradient ascent step}}
\EndForEach
\end{algorithmic}
\end{algorithm}
In line \ref{opt1} of the pseudocode, $\textrm{Breg}_F(x||y)$ denotes the usual Bregman divergence between the points $x$ and $y$ with respect to the convex function $F(\cdot),$ \emph{i.e.,}
\begin{eqnarray*}
	\textrm{Breg}_F(x||y) = F(x)-F(y)- \langle \nabla F(y), x-y\rangle .
\end{eqnarray*}

%\subsection{Simulation details and Additional Numerical Results}  \label{sim-addl}
\section{Efficient implementation of the optimization module} \label{opt_implementation}

To speed up the simulation, we implemented a custom-made optimizer for the optimization steps \ref{opt1} and \ref{opt2} involved in the \BQ algorithm in the bandit-feedback setting. For this, we directly solved the KKT optimality condition, where we computed the optimal KKT multiplier by using the classic Newton-Raphson root-finding algorithm. This empirically resulted in about \emph{two orders} of magnitude speed-up compared to using standard convex optimization packages such as \texttt{CVX} \citep{grant2011cvx}.  

%\cmt{Report the run-time and other details here}. 

Let $\bm{r} \in \mathbb{R}^N$ be a given $N$-dimensional real vector. After some simple algebraic manipulations, both the optimization problems in steps \ref{opt1} and \ref{opt2} of Algorithm \ref{fair-MAB-bandit-info} %for a generic input parameter vector $\bm{r}$ 
can be expressed in the following form: 

\begin{eqnarray} \label{obj_fun}
\texttt{OPT}(\bm{r}):~	\max \sum_{i=1}^N \log x_i + \langle \bm{r}, \bm{x} \rangle
\end{eqnarray}
Subject to,
\begin{eqnarray} \label{constr}
	\sum_i x_i=1,~ x_i \geq 0, ~\forall i \in [N]. 
\end{eqnarray}

Since the objective function \eqref{obj_fun} is strictly concave, and the constraint \eqref{constr} is linear, using the KKT condition, a probability vector $\bm{x}^*$ is an optimal point for the above problem if and only if there exists a real number $\mu \in \mathbb{R}$ s.t. 
\begin{eqnarray} \label{kkt1}
	\frac{1}{x_i^*} + r_i + \mu =0 \implies x_i^*= - (r_i+\mu)^{-1}, ~\forall i,
\end{eqnarray} 
where $\bm{x}^*$ satisfies the feasibility condition \eqref{constr}. For the non-negativity constraint on $\bm{x}^*$, we must have: 
\begin{eqnarray*}
	r_i+\mu < 0 \implies \mu < -\max_i r_i.
\end{eqnarray*}
Finally, we require that 
\begin{eqnarray*}
	\sum_i x_i^* =1.
\end{eqnarray*}
\emph{i.e.,}
\begin{eqnarray} \label{newton-raphson}
	\sum_i \frac{1}{r_i+\mu}-1=0.
\end{eqnarray}
We now use the Newton-Raphson method for solving \eqref{newton-raphson} starting from $\mu^{(0)}=  -\max_i r_i -1.$  The algorithm is given below:

  \begin{algorithm}
\caption{Custom optimizer for the problem \texttt{OPT} ($\bm{r}$)}
\label{optimizer}
\begin{algorithmic}[1]
\State \algorithmicrequire{ $\bm{r},$  $\texttt{tolerance}\gets 10^{-8}.$ }
\State $\mu\gets  -\max_i r_i -1, \texttt{error}\gets 1.$
\While{$\texttt{error}>\texttt{tolerance}$}
 \begin{eqnarray*}
	\mu \gets \mu + \frac{\sum_i \frac{1}{r_i+\mu}-1}{\sum_i \frac{1}{(r_i+\mu)^2}}.
\end{eqnarray*}
$\texttt{error} \gets |\sum_i \frac{1}{r_i+\mu}-1|.$
\EndWhile
\State $x_i^* \gets -(r_i+\mu)^{-1}, ~\forall i.$
\State Return $\bm{x}^*.$
\end{algorithmic}
\end{algorithm}
% The projection-step on to in the full information setting is implemented using the 

\section{Additional numerical results} \label{addl_sim}
\subsection{Comparison with an oracle policy}\label{addl_sim1}
\begin{figure}[h!]
	\centering
	\includegraphics[scale=0.4]{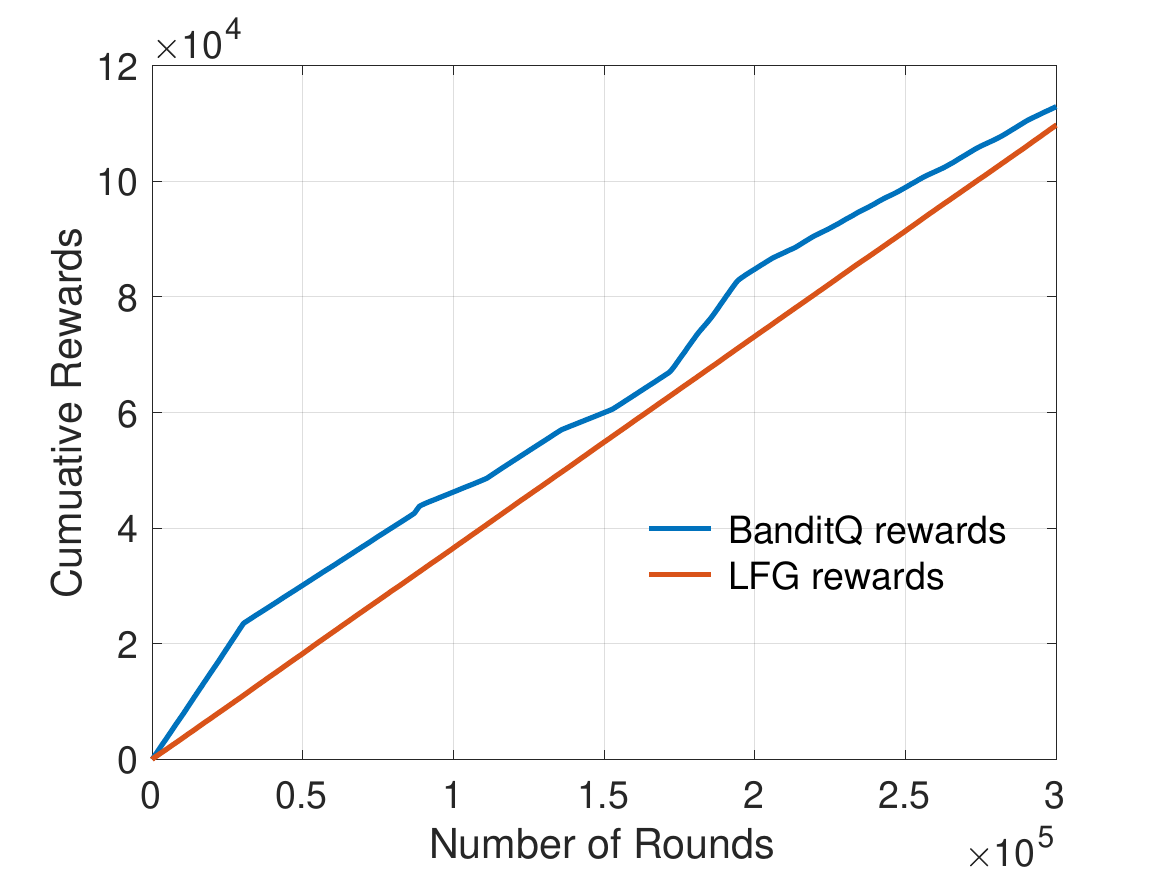}
	\caption{Comparison of reward accrued by the \BQ policy and the Oracle \texttt{LFG} policy ($\eta = 100$)}
	\label{rew-comp}
\end{figure}
In this section, we compare the performance of the \BQ policy with an \emph{Oracle} policy that knows the optimal fraction of pulls of each arm to satisfy the required reward rate constraints. With the given mean reward $\bm{\mu}$ and the required reward rate vector $\bm{\lambda}$, the optimal fraction of pulls can be easily computed to be $f_1=\nicefrac{\lambda_1}{\mu_1}=\nicefrac{1}{2}, f_2=\nicefrac{\lambda_2}{\mu_2}=\nicefrac{1}{3}, f_3=0, f_4=1-(\nicefrac{1}{2}+\nicefrac{1}{3})=\nicefrac{1}{6}, f_5=0.$ In the above computation, we have used the fact that Arm \#4 is the most rewarding arm. We emphasize that the oracle policy should have \emph{exact} knowledge of the mean reward vector $\bm{\mu}$ - a non-zero error in the value of the reward vector either leads to not achieving the target rates or having a linear regret or both.  

Note that the online policy proposed by \citet{patil2021achieving} \emph{cannot} be used with the above profile of fraction of pulls as their policy requires the required fraction of each arm to be at most $\nicefrac{1}{N-1} = \nicefrac{1}{4}.$ Hence, we use the UCB-based policy proposed by \citet{li2019combinatorial}, called \emph{Learning with Fairness Guarantee} (\textsc{LFG}), as the benchmark. \textsc{LFG} uses queue variables to balance meeting the target fraction of pulls and achieving the small regret. However, as stated in \citet[Theorem 2]{li2019combinatorial}, the best-known regret bound of the \textsc{LFG} policy increases linearly with time.

\paragraph{Observation:} From Figure \ref{rew-comp}, we see that the proposed \BQ policy yields strictly better cumulative rewards compared to the oracle \textsc{LFG} policy that knows the optimal fraction of arm pulls to meet the given reward rate constraints. This result can be attributed to the fact that the \BQ policy directly takes into account the reward realizations through the queue evolutions, whereas the Oracle \textsc{LFG} policy works based on the expected rewards only.  

\subsection{Large-scale experiments with $N=1000$ arms} \label{addl_sim2}

\begin{figure*}[t]
  \centering
  \begin{minipage}[b]{0.3\linewidth}
   \centering
    \includegraphics[width=\linewidth]{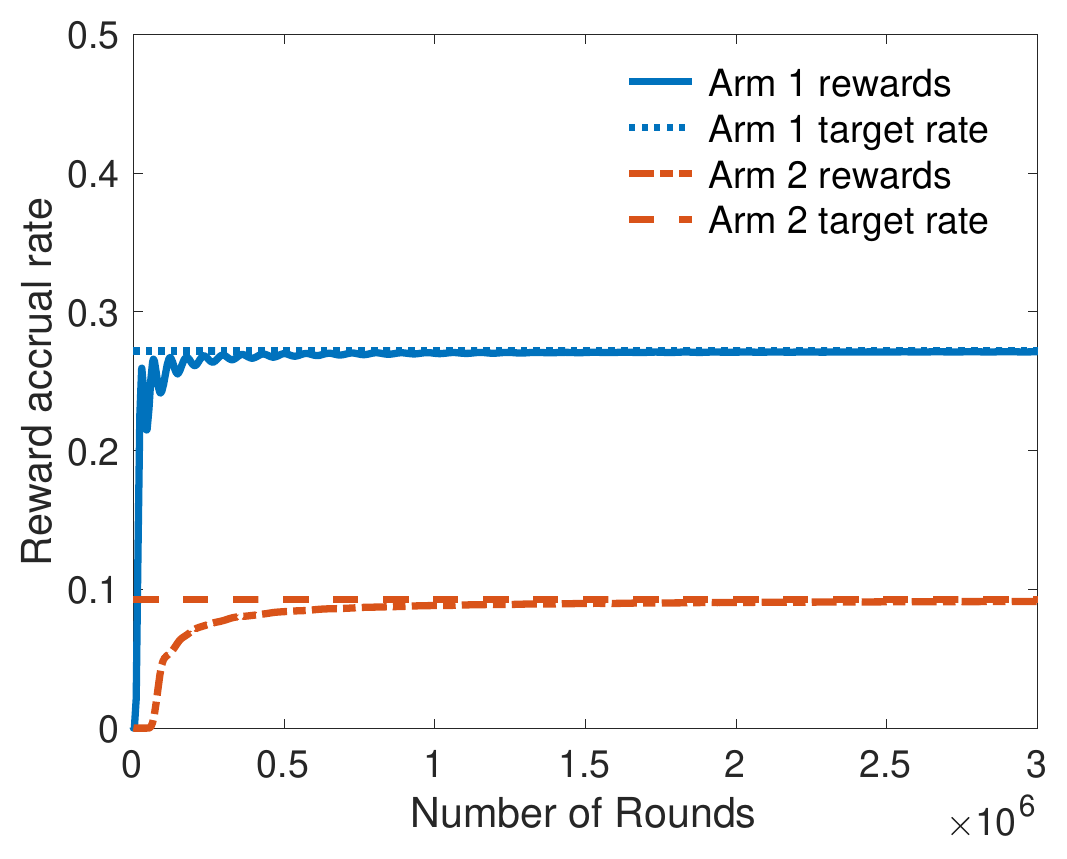}
   \caption{\small{Reward accrual rates in the full-information setting}}
   \label{rew_full}
  \end{minipage}
   \begin{minipage}[b]{0.3\linewidth}
   \centering
    \includegraphics[width=\linewidth]{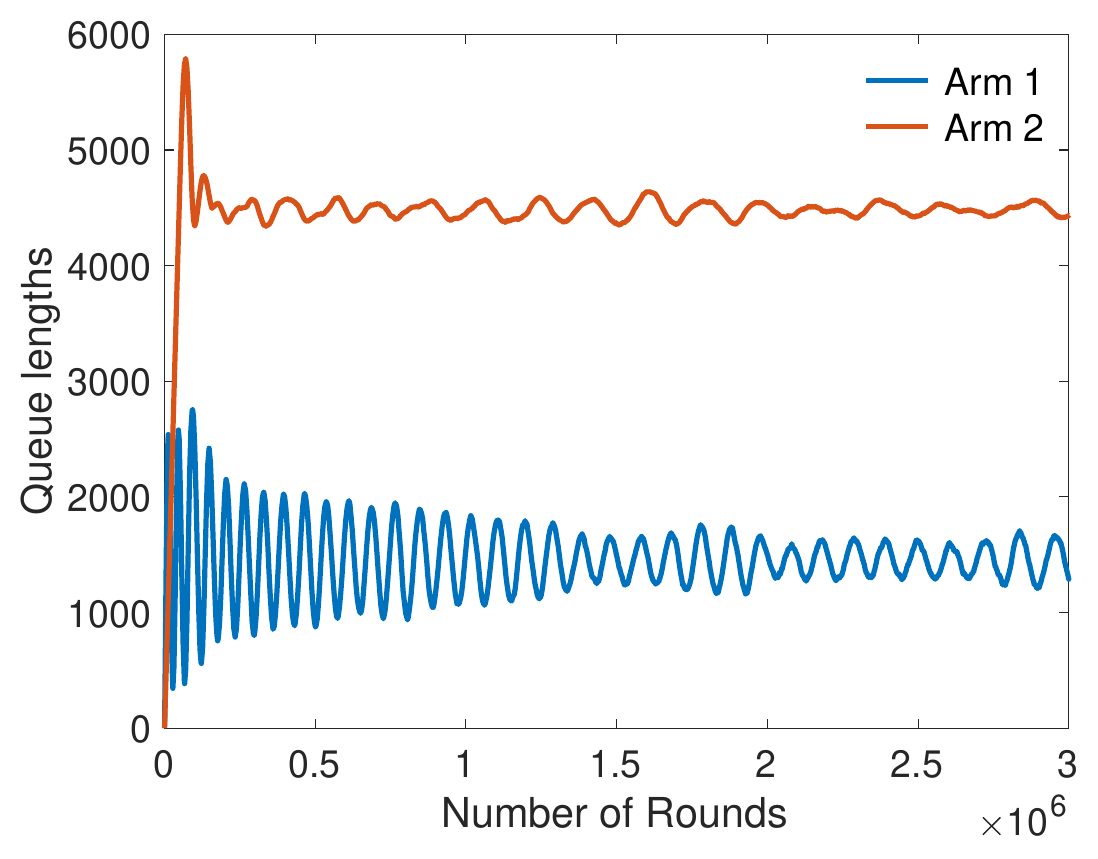}
   \caption{\small{Queue lengths in the full-information setting}}
   \label{q_full}
  \end{minipage}
   \begin{minipage}[b]{0.3\linewidth}
   \centering
    \includegraphics[width=\linewidth]{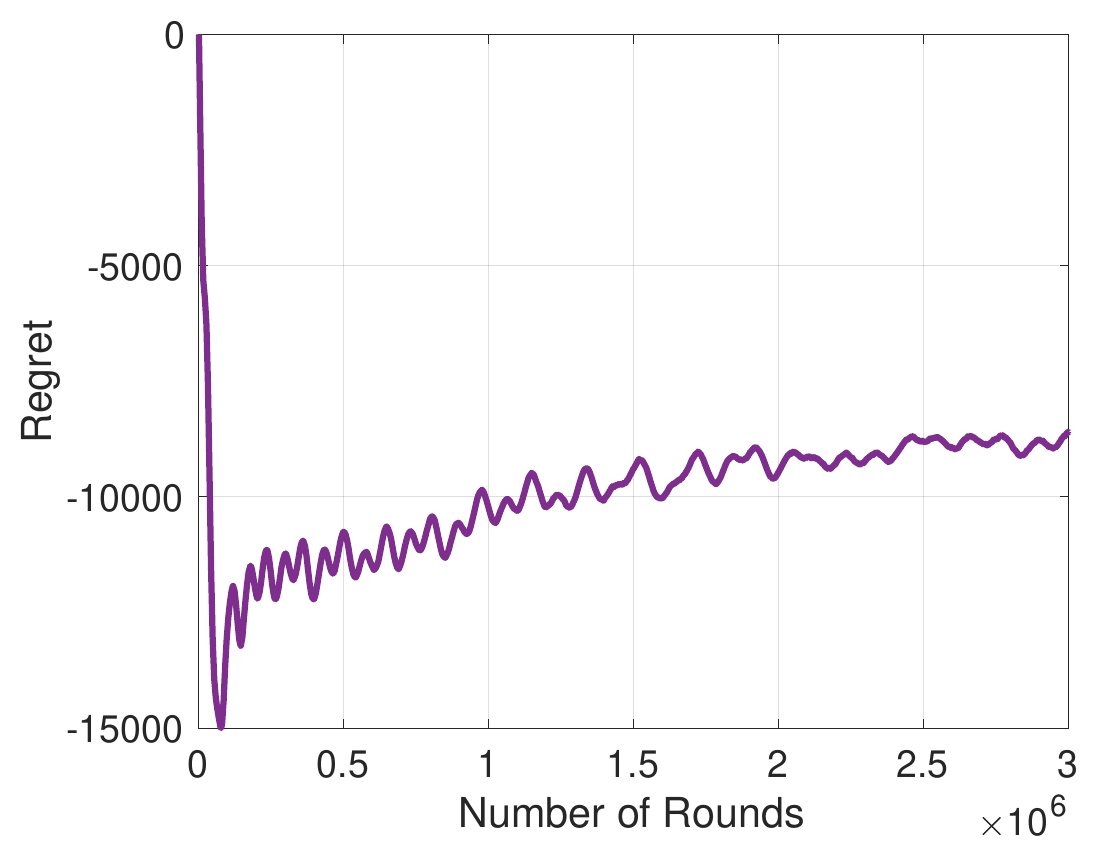}
   \caption{\small{Regret of \BQ in the full-information setting}}
   \label{reg_full}
  \end{minipage}
  \begin{minipage}[b]{0.3\linewidth}
   \centering
    \includegraphics[width=\linewidth]{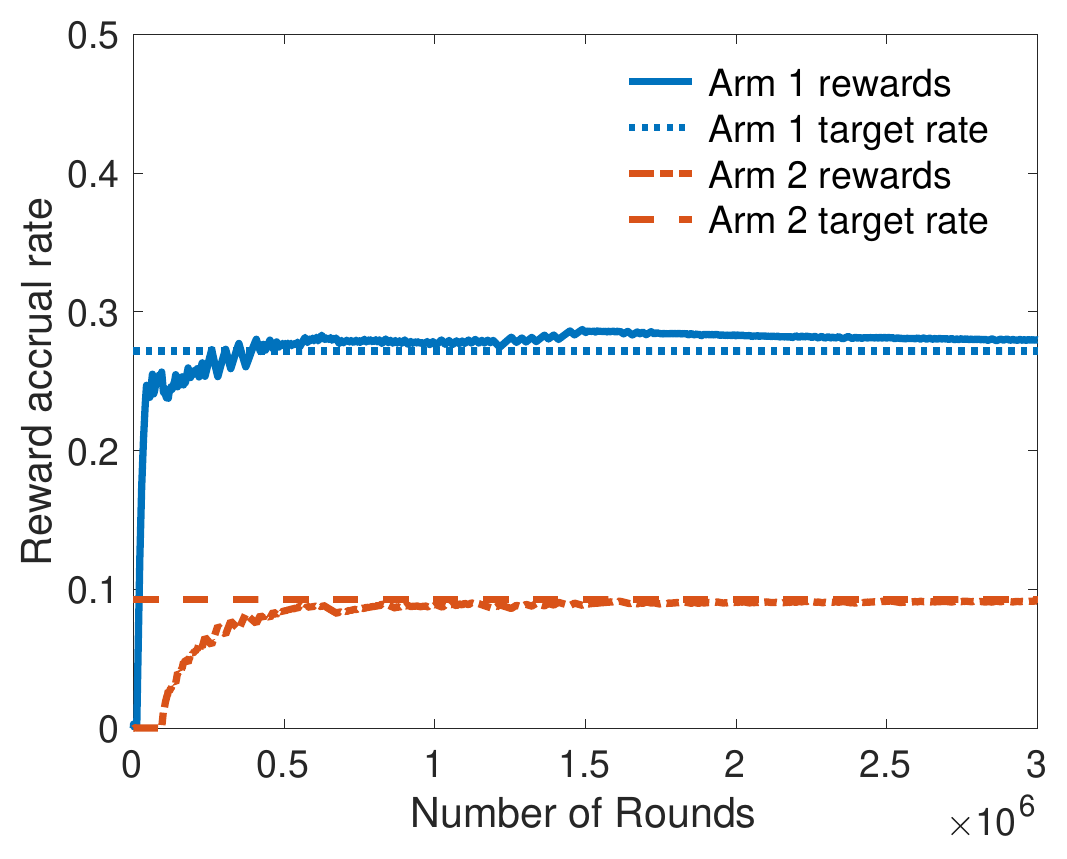}
   \caption{\small{Reward accrual rates in the bandit feedback}}
   \label{rew_bf}
  \end{minipage}
  %\hfill
  \begin{minipage}[b]{0.3\linewidth}
   \centering
    \includegraphics[width=\linewidth]{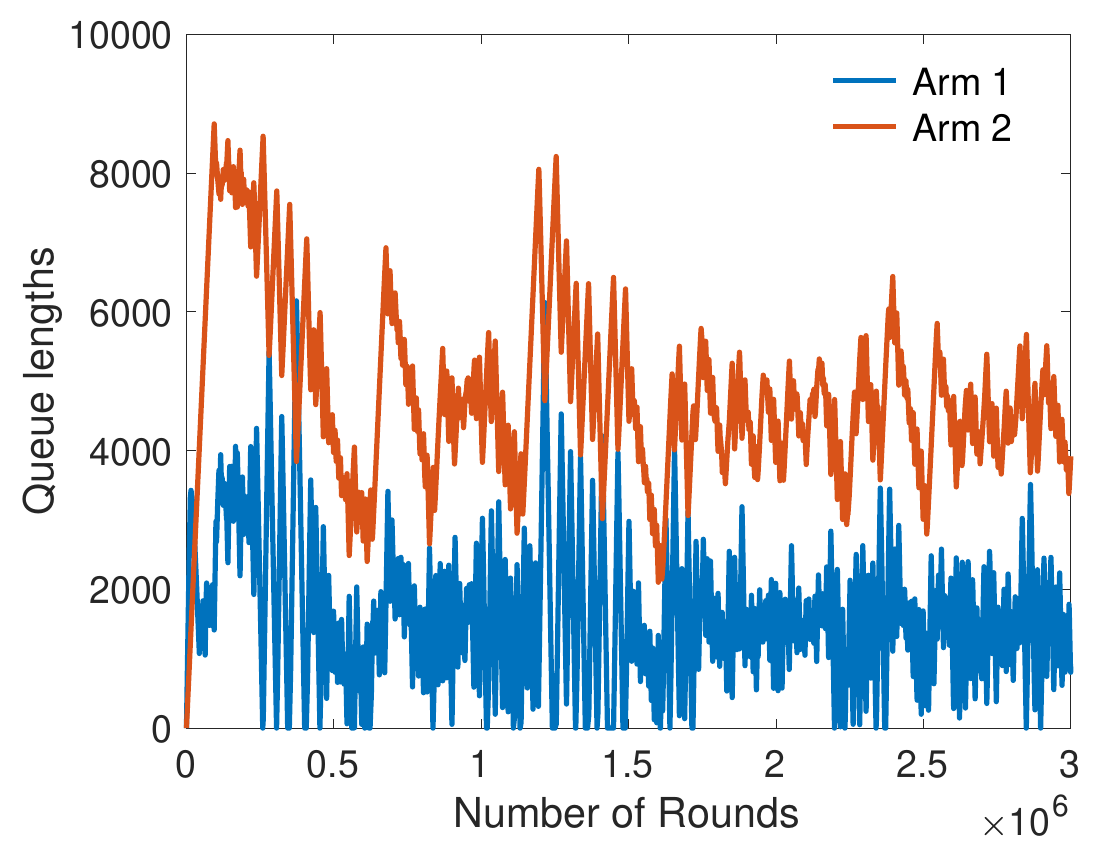}
   \caption{\small{Queue lengths in the bandit feedback setting}}
   \label{q_bf}
  \end{minipage}
 %\hfill
   \begin{minipage}[b]{0.3\linewidth}
   \centering
    \includegraphics[width=\linewidth]{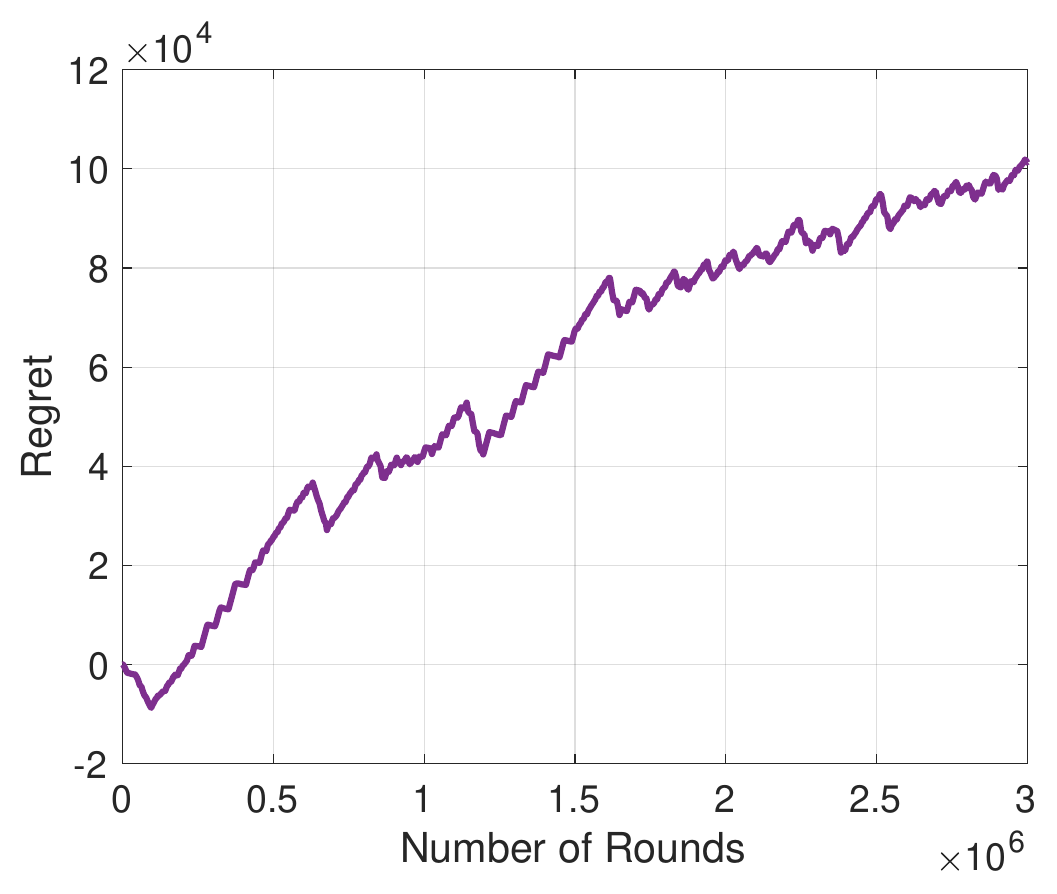}
   \caption{\small{Regret of \BQ in the bandit feedback setting}}
   \label{reg_bf}
  \end{minipage}
\end{figure*}
Figure [8-13] show the performance of the BanditQ policy with $N=1000$ arms in both full and bandit information setting. The mean reward for each arm is sampled uniformly at random from the interval $[0,1].$ As before, we consider two protected arms - arm 1 and arm 2 and set $\lambda_1= \mu_1/2, \lambda_2=\mu_2/3.$ The plots show that even for a large instance, the \BQ policy continues to perform satisfactorily in terms of both regret and achieving the target rates.

\end{document}